\pdfoutput=1
\documentclass[11pt]{article}

\usepackage[]{EMNLP2022}

\usepackage{times}
\usepackage{latexsym}

\usepackage[T1]{fontenc}
\usepackage[utf8]{inputenc}

\usepackage{microtype}

\usepackage{inconsolata}

\usepackage{multirow}
\usepackage{booktabs}
\usepackage{subfigure}
\usepackage{graphicx}
\usepackage{amsmath}
\usepackage{makecell}

\usepackage{enumitem}
\usepackage{tcolorbox}

\title{Benchmarking Language Models for Code Syntax Understanding}

\newcommand{\affmark}[1][*]{\textsuperscript{#1}}
\newcommand{\email}[1]{\texttt{\href{mailto:#1}{#1}}}

\author{%
Da Shen\affmark[1], Xinyun Chen\affmark[2]$^{\dagger}$, Chenguang Wang\affmark[3]$^{\dagger}$, Koushik Sen\affmark[4], Dawn Song\affmark[4]\\
\affmark[1]University of Maryland, College Park, \affmark[2]Google Research, Brain Team\\
\affmark[3]Washington University in St. Louis, \affmark[4]University of California, Berkeley\\
\email{dashen@terpmail.umd.edu}, \email{xinyunchen@google.com}, \email{chenguangwang@wustl.edu},\\
\email{\{ksen,dawnsong\}@cs.berkeley.edu}
}

\begin{document}
\newcommand{\eat}[1]{}
\def\ours{\texttt{CodeSyntax}}
\newif\ifsubmit
\submitfalse

\newcommand{\maybetodo}[1]{\ifsubmit{}\else{#1}\fi}

\ifsubmit
\newcommand{\xinyun}[1]{{\maybetodo{\color{red}}}}
\else
\newcommand{\xinyun}[1]{{\maybetodo{\color{red}[Xinyun: #1]}}}
\fi

\newcommand{\todo}[1]{{\color{red} TODO: {#1}}}

\newcommand{\boldgreen}[1]{{\textbf{\color{green!70!black}{#1}}}}
\newcommand{\boldred}[1]{{\textbf{\color{red}{#1}}}}
\newcommand{\boldblue}[1]{{\textbf{\color{blue}{#1}}}}

\hypersetup{urlcolor=black}
\maketitle

\def\thefootnote{$^\dagger$}\footnotetext{Corresponding authors.}
\def\thefootnote{\arabic{footnote}}

\hypersetup{urlcolor=darkblue}
\begin{abstract}
Pre-trained language models have demonstrated impressive performance in both natural language processing and program understanding, which represent the input as a token sequence without explicitly modeling its structure. Some prior works show that pre-trained language models can capture the syntactic rules of natural languages without finetuning on syntax understanding tasks. However, there is limited understanding of how well pre-trained models understand the code structure so far. In this work, we perform the first thorough benchmarking of the state-of-the-art pre-trained models for identifying the syntactic structures of programs. Specifically, we introduce \ours{}, a large-scale dataset of programs annotated with the syntactic relationships in their corresponding abstract syntax trees. Our key observation is that existing language models pretrained on code still lack the understanding of code syntax. In fact, these pre-trained programming language models fail to match the performance of simple baselines based on positional offsets and keywords. 
We also present a natural language benchmark to highlight the differences between natural languages and programming languages in terms of syntactic structure understanding. Our findings point out key limitations of existing pre-training methods for programming languages, and suggest the importance of modeling code syntactic structures.\footnote{Our code and dataset are available at \url{https://github.com/dashends/CodeSyntax}.}
\end{abstract}

\section{Introduction}

\begin{figure}[!ht]
\centering
\includegraphics[width=0.9\linewidth]{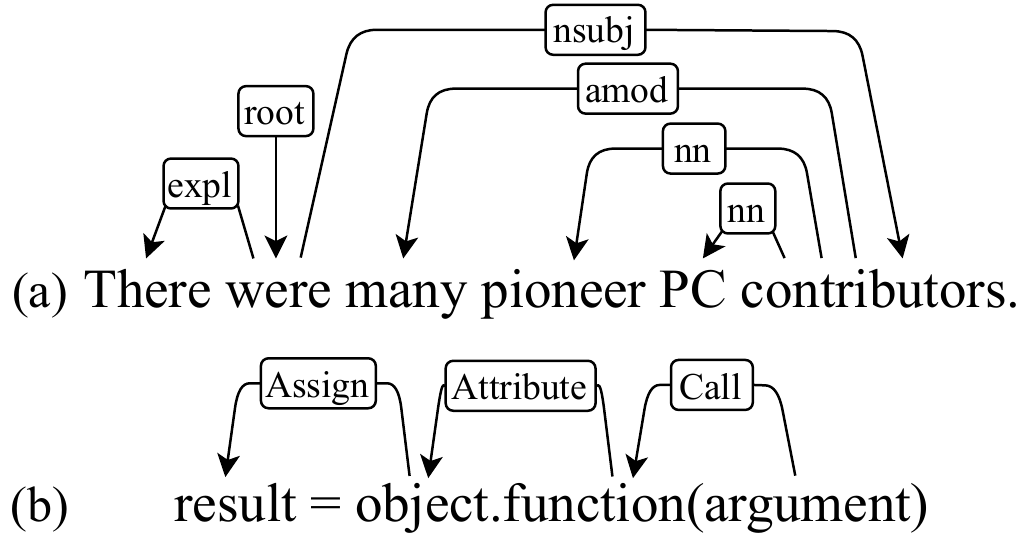}

\caption{Examples of syntactic relations for (a) natural languages (NL) and (b) programming languages (PL). Each relation is represented by an arrow. The relations in PL represent the syntax of code in a way similar to those in NL.}
\label{fig:nl_pl_dep}
\end{figure}

\begin{figure}[!htbp]
\centering

% \includegraphics[scale=0.4]{figures/preview_NL.pdf}
% \vspace{-0.8em}
% \includegraphics[scale=0.4]{figures/preview_PL.pdf}
\includegraphics[scale=0.36]{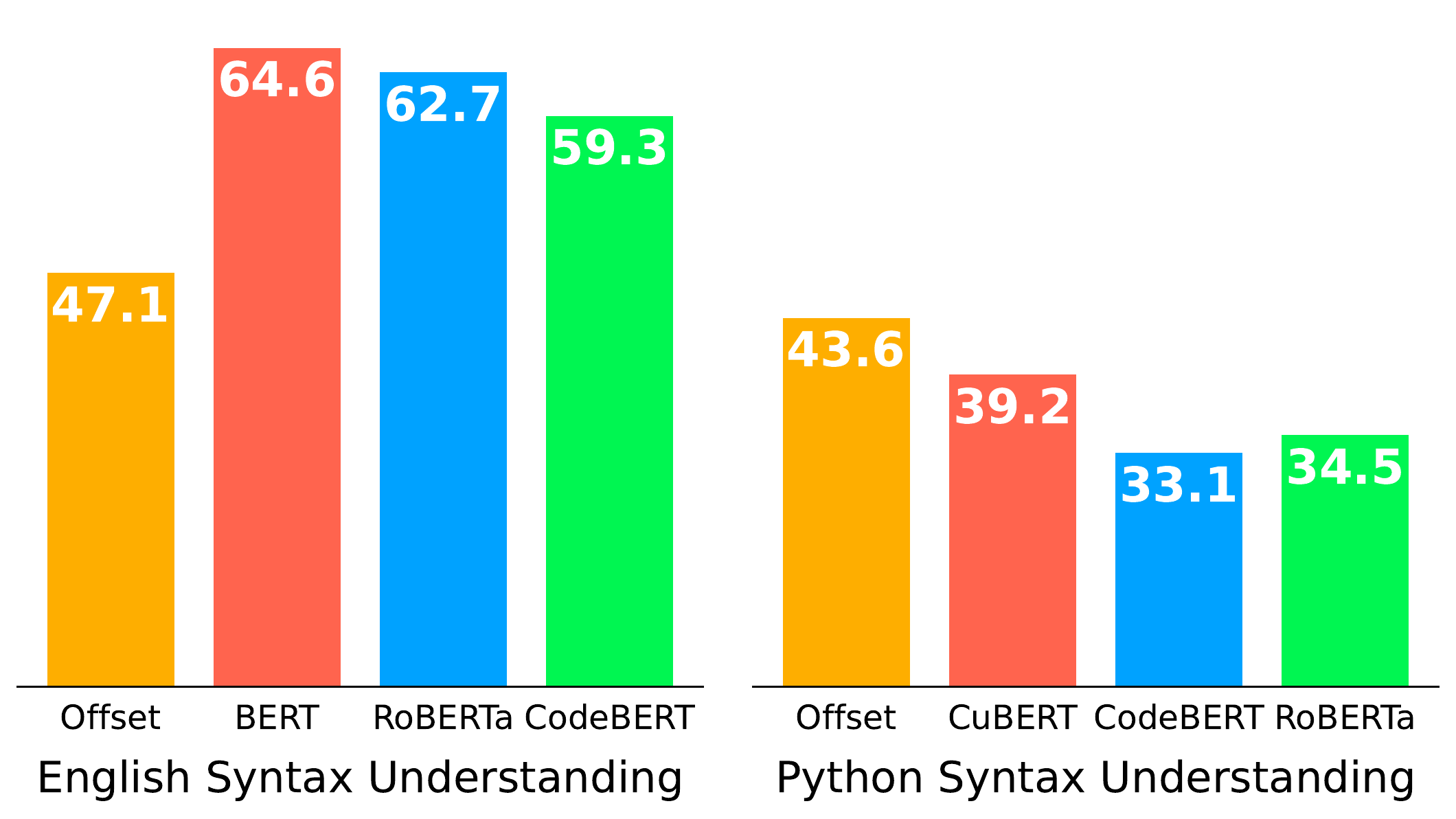}

\caption{A preview of the model performance comparison on NL and PL syntax understanding tasks. Pre-trained models capture NL syntax relatively well, but perform worse in understanding PL syntax. The \emph{Offset} baseline picks the token using a fixed positional offset.  
We use BERT-large and RoBERTa-base configurations (corresponding to the configurations of CuBERT and CodeBERT). The plot shows top-1 scores. See Tables~\ref{tab:main-res-PL} and~\ref{tab:main-res-NL} for the full results.}
\label{tab:res-preview}

\end{figure}

Large-scale pre-training of language models has become the de-facto paradigm for a variety of natural language processing tasks. Furthermore, recent studies show that models pre-trained on a massive amount of code also achieve competitive performance on many tasks, e.g., code generation and code classification. These tasks are closely related to natural language (NL) tasks in their problem formulation. Nowadays, the common practice for solving these coding tasks is to utilize the language model architectures and training schemes that are originally designed for NL. The design principle of these neural language models is significantly different from the classic rule-based program generation systems. Specifically, neural language models take the program as a token sequence, while classic program generation systems utilize the language grammar and code structure. Despite the advanced performance of pre-trained language models on code understanding tasks, what these models have learned from the code corpus remains unclear.

In this work, we investigate whether large-scale pre-training is all we need for code representation learning. In particular, we conduct the first systematic study to analyze how the pre-trained language models understand the syntactic structures of programs. To this end, we introduce \ours{}, a large-scale benchmark consisting of programs annotated with the syntactic relationships between different tokens. The ground truth syntactic relationships are extracted from edges in the abstract syntax trees (AST) of the programs. Figure~\ref{fig:nl_pl_dep} shows some examples. These syntactic relations are function-wise similar to dependency relations for NL, where prior work has demonstrated that the attention heads of pre-trained language models can help to identify NL relation types~\cite{clark2019does,raganato2018analysis}. To measure how well the pre-trained language models capture the code syntactic structures, we adopt the approach to the PL domain. We focus on investigating the zero-shot capability of existing pre-training methods in our experiments, and we evaluate these pre-trained models without finetuning them on our benchmark.

% \begin{table}[ht]
% \centering
% \begin{tabular}{@{}lll@{}}
% \toprule
% Language & Model & Score \\
% \midrule
% \multirow{5}{*}{Python} 
%     & Offset & 43.6 \\
%     & Combined & \textbf{49.4} \\
%     \cline{2-3}
%     & CuBERT & 39.2 \\
%     & CodeBERT & 33.1 \\
%     & RoBERTa & 34.5 \\\midrule
% \multicolumn{2}{l}{Diff (Model - Baseline)} & -10.2
% \\ \midrule
% \multirow{3}{*}{English} 
%     & Offset & 47.1 \\
%     \cline{2-3}
%     & BERT-large & \textbf{64.6} \\
%     & RoBERTa-base & 62.7  \\
%     & CodeBERT & 59.3 \\
%     \midrule
% \multicolumn{2}{l}{Diff (Model - Baseline)} &17.5 \\
% \bottomrule
% \end{tabular}
% \caption{A preview of the model performance comparison on programming language and natural language syntax understanding. The baseline~\emph{Offset} always picks the token with a fixed positional offset, and~\emph{Combined} picks the best option from a fixed positional offset or a fixed keyword. We present the top-1 scores, and the score differences are calculated as the best attention score - best baseline score for each language, where a positive value indicates that the language model surpasses the baseline performance. See Tables~\ref{tab:main-res-PL} and~\ref{tab:main-res-NL} for the full results.}
% \label{tab:res-preview}
% \end{table}

We evaluate the state-of-the-art pre-trained language models for code representation learning, including CuBERT~\cite{kanade2020learning} and CodeBERT~\cite{feng2020codebert}. A common characteristic of these models is that they share the same Transformer-based architectural design as NL models~\cite{vaswani2017attention,devlin2019bert}. This allows us to directly compare their performance in capturing the syntax structure. We present a preview of our key results in Figure~\ref{tab:res-preview}. Our main observation is that pre-training is insufficient for learning the syntactic relations in code. First, we find that the models pre-trained on code do not always outperform models pre-trained on NL corpus alone. Surprisingly, compared to CodeBERT which is trained on both text and code corpora, RoBERTa achieves better performance without training on any code with identical model architecture. This indicates that pre-training on programs as token sequences does not help learn the syntactic relations. On the contrary, without dependency relations, pre-training still enables language models to understand the NL syntax to some extent.

Moreover, for code syntax understanding, the pre-trained models even perform worse than simple baselines that pick the tokens with a fixed offset. For example, always selecting the (p+2)-th token as the p-th token's dependency yields higher accuracy than any attention head for several relation types. On the other hand, the same model architectures pre-trained on text corpora achieve decent accuracy in identifying the dependency relations in the NL domain, where the performance of the same simple baselines is far behind.

Our analysis reveals several key differences between NL and PL that lead to different capabilities of understanding the syntax for pre-trained models. First, programs are more structured than NL sentences. Programs usually contain hierarchical structures representing long-term dependencies between code tokens. Consequently, a large number of syntactic relation types are between distant tokens, which can be difficult to recognize for attention heads. On the contrary, the dependency relations in NL sentences mostly connect nearby token pairs, and in this case the attention heads are more capable of identifying the correct relations. Meanwhile, language models are good at recognizing keyword-based relations, such as picking the corresponding~\emph{else} keyword for an~\emph{if} token. Interestingly, we find that the inclusion of tokens such as newlines and semicolons notably affects the performance in the code domain.

Our findings suggest that existing pre-trained models perform quite differently in PL and NL domains in terms of the ability to understand syntax. Thus, directly applying training paradigms developed for NL could be suboptimal for program learning, and we consider designing better approaches to model the code structure as future work.
\section{\ours{}: Benchmarking Code Syntax Understanding}

\begin{table*}[ht]
\setlength\tabcolsep{5pt}
\centering
\scalebox{0.9}{
\begin{tabular}{lrrp{4.6cm}ll}
\toprule
Relation & \multicolumn{2}{c}{Count} & \multirow{2}{*}{Explanation}         & \multicolumn{2}{l}{~~~~~~~~Code Example}  \\
 head$\rightarrow$dependent & Python      & Java        &      & Python           & Java            \\
 \midrule
\makecell{Assign:\\target$\rightarrow$value} & 78,482 & 13,384 & Assigning a value to a target variable. & \textbf{\color{blue}target} = \textbf{\color{red}10} & int \textbf{\color{blue}target} = \textbf{\color{red}10}; \\
\hline
\makecell{Call:\\func$\rightarrow$args} & 110,949 & 50,890 & Calling a function with some arguments. & \textbf{\color{blue}function}(\textbf{\color{red}arg}) & \textbf{\color{blue}function}(\textbf{\color{red}arg});\\
\hline
\makecell{For:\\for$\rightarrow$body} & 8,704 & 1,864 & A for loop repeatedly executes the body block for some iterations. & \makecell{\textbf{\color{blue}for} target in iter:\\ ~~~~~\textbf{\color{red}body}}& \makecell{\textbf{\color{blue}for} (initializers; \\~~~~~test; updaters) \{\\ ~~~~~\textbf{\color{red}body;} \\ \}}\\
\hline
\makecell{If:\\if$\rightarrow$else} & 11,024 & 5,038 & An if statement conditionally executes a body based upon some criteria. The dependent is the \texttt{else} keyword. & \makecell{\textbf{\color{blue}if} condition:\\ ~~~~~body1 \\ \textbf{\color{red}else}: \\ ~~~~~body2}& \makecell{\textbf{\color{blue}if} (condition) \{\\ ~~~~~body1; \\  \} \textbf{\color{red}else} \{ \\ ~~~~~body2; \\\} }\\
\hline
\makecell{If:\\if$\rightarrow$body} & 34,250 & 22,392 & An if statement. The dependent is the body block. & \makecell{\textbf{\color{blue}if} condition:\\ ~~~~~\textbf{\color{red}body1} \\ else: \\ ~~~~~body2} & \makecell{\textbf{\color{blue}if} (condition) \{\\ ~~~~~\textbf{\color{red}body1;} \\  \} else \{ \\ ~~~~~body2; \\\} } \\
\hline
\makecell{If:\\body$\rightarrow$orelse} & 11,024 & 4,976 & An if statement. The head is the body block and the dependent is the body of the else block. & \makecell{{if} condition:\\ ~~~~~\textbf{\color{blue}body1} \\ else: \\ ~~~~~\textbf{\color{red} body2}} & \makecell{{if} (condition) \{\\ ~~~~~\textbf{\color{blue}body1;} \\  \} else \{ \\ ~~~~~\textbf{\color{red} body2;} \\\} } \\
\hline
\makecell{While:\\test$\rightarrow$body} & 743 & 975 & The while loop repeatedly executes the body block as long as the specified condition is true.& \makecell{while \textbf{\color{blue}condition}:\\ ~~~~~\textbf{\color{red}body}}& \makecell{while (\textbf{\color{blue}condition}) \{\\ ~~~~~\textbf{\color{red}body;} \\ \}} \\
\bottomrule
\end{tabular}
}
\caption{ Dataset statistics of selected relation types in \ours{}. For each relation type, we highlight the head and dependent nodes in the examples in bold, with the head in blue and the dependent in red. We defer the full statistics of all relation types to Table~\ref{tab:data_stats_full} in the appendix.}
\vspace{-1em}
\label{tab:data_stats}
\end{table*}

We construct the \ours{}  benchmark to evaluate the performance of language models on code syntax understanding. We focus on Python and Java languages, on which the publicly released model checkpoints of both CuBERT~\cite{kanade2020learning} and CodeBERT~\cite{feng2020codebert} are pre-trained. We obtain the code samples from CodeSearchNet~\cite{husain2019codesearchnet}, which is a large-scale dataset consisting of code in different programming languages. Its training set is also part of the pre-training data of CodeBERT, so we remove the data samples that are included in the pre-training data of either CuBERT or CodeBERT. Thus, none of the programs in \ours{} has been seen by CuBERT or CodeBERT in the pre-training phase. 

 In total, \ours{} contains 18,701 code samples annotated with 1,342,050 relation edges in 43 relation types for Python, and 13,711 code samples annotated with 864,411 relation edges in 39 relation types for Java. Each code sample is an entire function consisting of multiple statements, which is analogous to a paragraph in NL. Each relation corresponds to an edge in the program AST; specifically, we utilize the Python ast module~\cite{pythonast2021} and the Java org.eclipse.jdt.core.dom.ASTParser class~\cite{javaast2014} to parse a code sample into an AST. We present some examples of relation types in Table~\ref{tab:data_stats}, and we defer the description of all relation types to Table~\ref{tab:data_stats_full} in the appendix. More details about relation extraction are discussed in Appendix~\ref{app:data-construction}. Note that we can easily extend the dataset to cover more languages since the workflow for extracting relations is automated and AST parsers are available for most popular programming languages.

\begin{figure}[htb]
\centering
\vspace{-1em}
\subfigure[\ours{}.]{%
\includegraphics[scale=0.5]{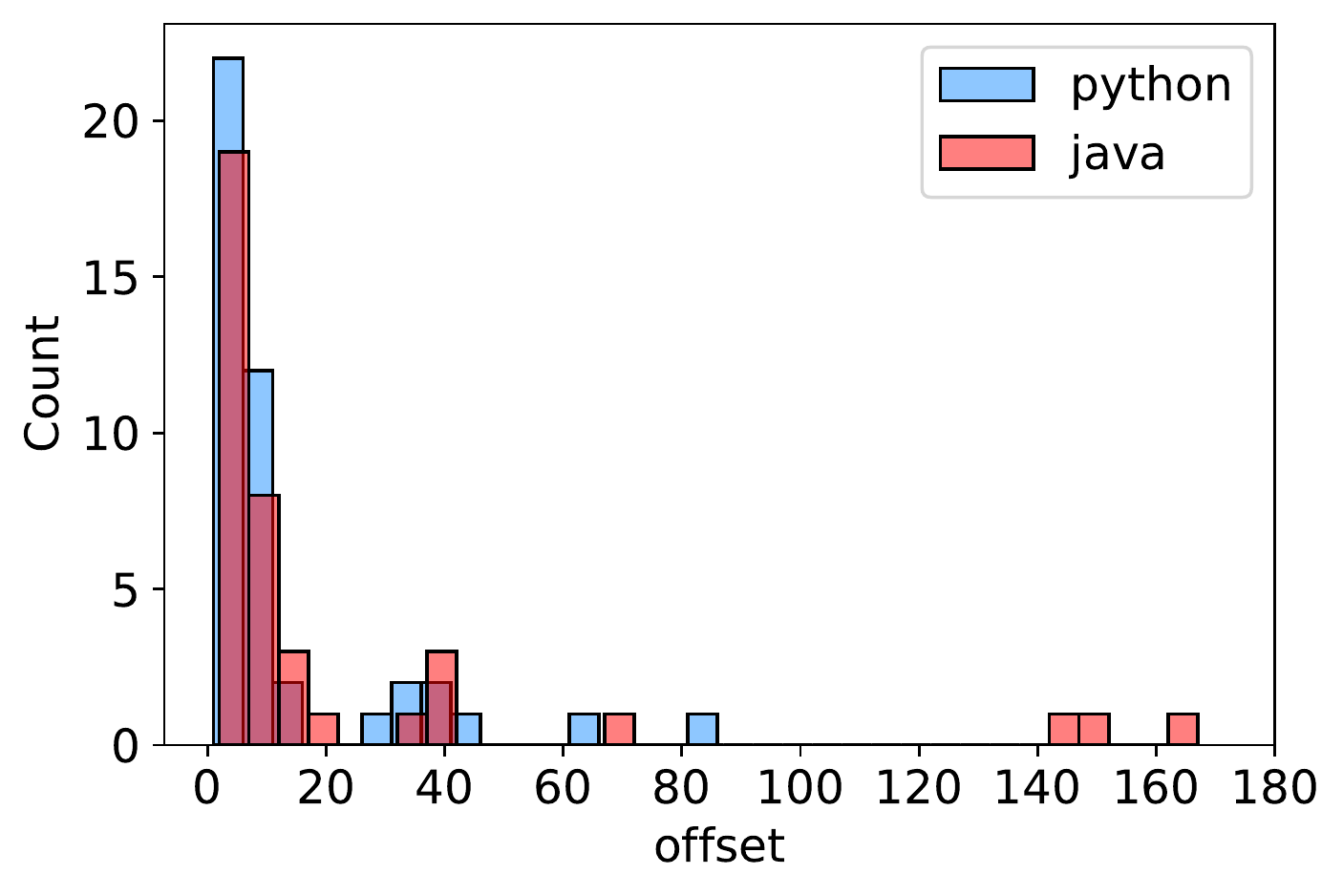}}
\vspace{-1em}
\subfigure[Natural language corpus.]{%
\includegraphics[scale=0.5]{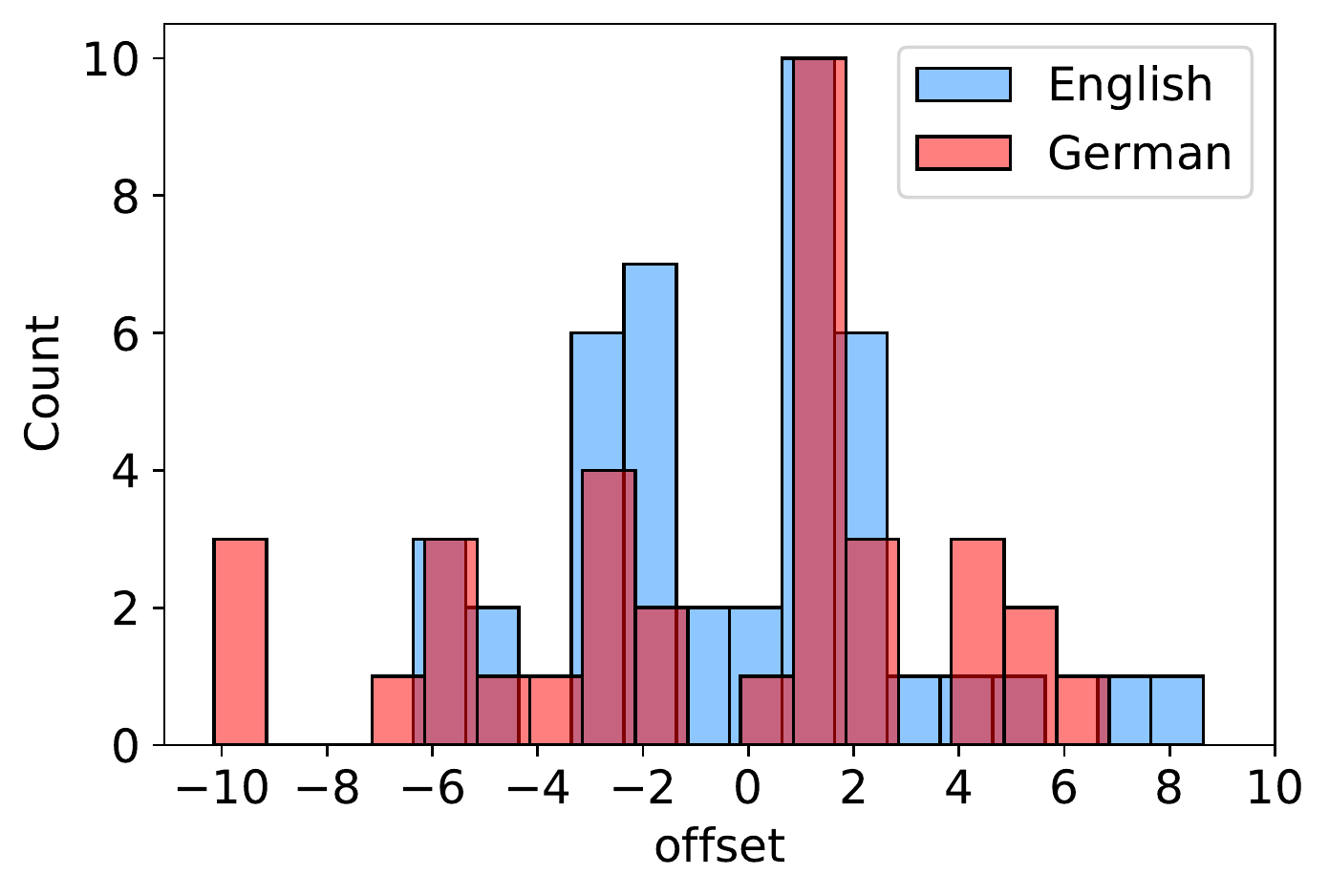}}
\caption{Offset distribution of relation types in (a) \ours{} and (b) NL corpus. The x axis is the average positional offset distance between heads and dependents for each relation. The y axis is the number of relations that has the average offset value. See Section~\ref{sec:exp-setup} for more details on the NL corpus.}
\label{fig:offset_dist}
\vspace{-1em}
\end{figure}

% \begin{figure}[ht]
% \centering
% \includegraphics[width=\linewidth]{figures/offset_distribution_start.pdf}
% \caption{Offset distribution of relation types in \ours{}. The x axis is the average positional offset distance between heads and dependents for each relation. The y axis is the number of relations that has the average offset value.}
% \label{fig:offset_dist_start}
% \end{figure}

% \begin{figure}[ht]
% \centering
% \includegraphics[width=\linewidth]{figures/offset_distribution_NL.pdf}
% \caption{Offset distribution of relation types in the natural language corpus. See Section~\ref{sec:exp-setup} for more details.}
% \label{fig:offset_dist_NL}
% \end{figure}

We observe several characteristics of relations in \ours{}. First, the keywords in PL play an important role in recognizing the code structure. Specifically, some relation types have fixed keywords as the edge nodes, such as the \texttt{If:if$\rightarrow$else} relation. Meanwhile, compared to the dependency relations in NL, the relation edges in the program AST tend to connect nodes that are much farther away from each other. As shown in Figure~\ref{fig:offset_dist}, the \hbox{average} offset between head and dependent nodes is no more than 10 for dependency relations in NL, while the average offset for a relation type can be more than 100 code tokens. Specifically, in \ours{}, there are 22 near dependency types whose average offsets are less than 10, and 12 far dependency types whose average offsets are above 10.

% Here we present all the relations considered in this paper, divided into several categories:

% \paragraph{Python Relations}
% \begin{itemize}
% \itemsep 0em 
%     \item Control Flow: If, For, While, Try, With
%     \item  Literals: Dict
%     \item   Expressions: BinOp, BoolOp, Compare, Call, IfExp, Attribute 
%     \item   Expr-Subscripting: Subscript, Slice
%     \item   Expr-Comprehensions: ListComp, GeneratorExp, DictComp, SetComp, comprehension
%     \item   Statements: Assign, AugAssign
%     \item   Vague: parent$\rightarrow$child
% \end{itemize}

% \paragraph{Java Relations}
% \begin{itemize}
% \itemsep 0em 
%     \item Control Flow: If, For, While, Try, Do, Switch
%     \item Expressions: InfixExpr, Call, IfExp, Attribute, InstanceofExpr 
%     \item Expr-Subscripting: Subscript
%     \item Statements: Assign (includes AugAssign)
%     \item   Vague: parent$\rightarrow$child
% \end{itemize}

\section{Evaluation Setup}
\label{sec:exp-setup}

Do pre-trained language models capture the code structure without direct supervision of the syntactic information? To investigate this question, we evaluate several pre-trained language models without finetuning, and compare their performance in understanding the syntax for NL and PL.

\paragraph{Natural language benchmark.} To compare the performance on \ours{} to NL syntax understanding, we construct the NL benchmark that includes English and German. Specifically, we use the English News Text Treebank: Penn Treebank Revised~\cite{ann2015} labeled with Stanford Dependencies~\cite{marie2008, DBLP:conf/coling/MarneffeM08}, and German Hamburg Dependency Treebank~\cite{DBLP:conf/lrec/FothKBM14} labeled with Universal Dependencies~\cite{DBLP:journals/coling/MarneffeMNZ21}. In total, the English dataset has 48,883 sentences, 43 relation types, and 1,147,526 relation edges; the German dataset has 18,459 sentences, 35 relation types, and 307,791 relation edges.

\paragraph{Attention probing approach.} Some prior works demonstrate that a Transformer architecture~\cite{vaswani2017attention} pre-trained on a text corpus, such as BERT~\cite{devlin2019bert}, contains attention heads that specialize in certain dependency relations in NL~\cite{raganato2018analysis,clark2019does}. Specifically, in the Transformer architecture, each vector $e_i$ for an input token is transformed into the query and key vectors $q_i$ and $k_i$ via some linear transformations, and the transformations vary among different attention heads. For the $i$-th token, the attention weight assigned to the $j$-th token is

\begin{equation*}
    \alpha_{i, j} = \frac{\exp(q_i^Tk_j)}{\sum_l\exp(q_i^Tk_l)}
\end{equation*}

The attention weight indicates how important the $j$-th token is with respect to the $i$-th token.

Typically, different attention heads learn different weights between input tokens. Therefore, to measure the correctness of recognizing a relation type \texttt{r}, for each edge \texttt{<h, t, r>} in the program AST where \texttt{h} is the head node and \texttt{t} is the dependent node, we enumerate all attention heads to compute the attention weight $\alpha_{h, t}$. If an attention head tends to assign high attention weights that connect the pair of tokens belonging to the relation type \texttt{r}, we consider the relation type to be captured. We defer more implementation details of attention map extraction to Appendix~\ref{app:attention-approach}.

\paragraph{Metrics.} We use the unlabeled attachment score (UAS) to measure the syntax understanding performance, and we consider top-k scores with different values of k. To compute top-k scores for language models, for each attention head, given the head token \texttt{h} in a relation edge \texttt{<h, t, r>}, we compute the attention weight over all tokens in the input code, and we consider the prediction to be correct if the attention weight over the dependent token \texttt{t} is among the top-k tokens with the highest attention weights. For each relation, we select the best-performing attention head and use its score as the model's score for that relation. We calculate a model's average score over all relations as the final score of the model.

In NL dependency parsing problems, the dependent node \texttt{t} usually corresponds to a single word. However, in PL, the dependent can be a block that contains multiple code tokens. For example, in the \texttt{If:if$\rightarrow$body} relation, the head is the keyword \texttt{if}, while the dependent is the entire body block. Therefore, we measure three metrics.  \emph{First-token metric} and \emph{last-token metric}: the prediction is deemed correct if it successfully predicts the first and last token of the dependent block, respectively; \emph{Any-token metric}: the prediction is considered correct if it can predict any token within the dependent block. While we agree that these are not perfect metrics and one single metric may be incomplete, we observe that our findings generally hold for all the three metrics we evaluated. Note that the first-token metric is stricter than the any-token metric by design. Unless otherwise specified, we report the top-k scores using the first-token metric by default.

\paragraph{Model architectures.} Table~\ref{tab:model_pairs} summarizes the models evaluated in this work. For language models over code, we consider CuBERT~\cite{kanade2020learning} and CodeBERT~\cite{feng2020codebert}, and we evaluate their released pre-trained checkpoints. Both of them are based on architectures initially designed for NL. Specifically, CuBERT utilizes the BERT~\cite{devlin2019bert} architecture, and CodeBERT~\cite{feng2020codebert} utilizes the RoBERTa~\cite{DBLP:journals/corr/abs-1907-11692} architecture. For NL models, we also evaluate multilingual variants of BERT and RoBERTa on the German dataset, i.e., Multilingual BERT~\cite{DBLP:conf/acl/PiresSG19} and XLM-RoBERTa~\cite{DBLP:conf/acl/ConneauKGCWGGOZ20}. Both of the two code language models are cased, so we also evaluate the cased versions of the NL models.

\begin{table}[ht]
\centering
\begin{tabular}{|l|l|}
    \hline
    Programming Languages & Natural Languages \\\hline
    \multirow{2}{*}{CuBERT} 
        & BERT  \\\cline{2-2}
        & Multilingual BERT \\
    \hline
    \multirow{2}{*}{CodeBERT} 
        & RoBERTa  \\\cline{2-2}
        & XLM-RoBERTa  \\
    \hline
\end{tabular}
\caption{Model architectures evaluated on PL and NL benchmarks. Models in the same row share the same architecture, but are pre-trained on different corpora.}
\label{tab:model_pairs}
\vspace{-1em}
\end{table}

\paragraph{Baselines.} To examine how well the attention performs through comparisons, we design a simple offset baseline and a simple keyword baseline. The offset baseline with an offset value of $i$ always selects the token after $i$ positions of the input token as its prediction when $i>0$, and selects $i$ positions before the input token when $i<0$. The keyword baseline with a keyword of $key$ always predicts the next $key$ token as its prediction. In our experiments, we evaluate offset baselines with each possible offset value between 0 and 512 for PL, and -512 to 512 for NL. We use all Python and Java keywords for the keyword baselines on Python and Java datasets respectively, including tokens such as \texttt{if}, \texttt{for}, \texttt{in}, etc. To evaluate the top-k scores for baselines where $k \geq 2$, we combine k 
simple baselines with different offset (keyword) values to give k predictions. To select k offset (keyword) values, we repeatedly and greedily include the next value that yields the highest performance increase for the relation type under consideration.

\eat{\paragraph{Code syntactic relation identification.} Given a programming language $\mathcal{L}$ and a program \texttt{P} $\in \mathcal{L}$, we view its abstract syntax tree \texttt{T} as a collection of edges \texttt{<h, t, r>}, where \texttt{h} and \texttt{t} are the connected nodes (i.e., tokens), and \texttt{r} indicates a syntactic relation type. For example, in a Python code \texttt{l = len(a)}, the edge \texttt{<len, a, Call>} means that tokens \texttt{len} and \texttt{a} are a function call with its corresponding argument. Given a relation type \texttt{r}, the code syntactic relation identification problem aims to identify the token \texttt{t} for any token \texttt{h} such that \texttt{<h, t, r>} $\in$ \texttt{T}.}

\section{Experiments}
\label{sec:exp}

In this section, we present the results of pre-trained language models for both PL and NL syntax understanding tasks, and discuss the key observations that distinguish PL from NL. 

\subsection{Main Results}
\label{sec:res}

\begin{table}[htbp]
\scalebox{0.85}{
\begin{tabular}{@{}llllll@{}}
\toprule
\multirow{2}{*}{Language} & \multirow{2}{*}{Model} & \multicolumn{4}{l}{Top-k Score} \\
                  &                   &   k=1  &    k=3 & k=10    & k=20     \\ \midrule
\multirow{5}{*}{Python} 
    & Offset & 43.6 & 63.7 & 87.3 & 94.9 \\
    & Keyword & 15.7 & 21.9 & 23.6 & 23.8 \\  
    & Combined & \textbf{49.4} & \textbf{69.7} & \textbf{90.1} & \textbf{96.3} \\
    \cline{2-6}
    & CuBERT & 39.2 & 58.4 & 81.3 & 91.4\\
    & CodeBERT & 33.1 & 51.8 & 78.6 & 89.2 \\
    & RoBERTa & 34.5 & 56.9 & 82.5 & 91.3 \\\midrule
\multicolumn{2}{l}{Diff (Model - Baseline)} & -10.2 & -11.3 & -8.8 & -4.9  
\\ \midrule
\multirow{5}{*}{Java} 
   & Offset & 52.7 & 71.5 & 87.1 & 94.3    \\
    & Keyword & 22.4 & 27.3 & 30.2 & 30.6  \\ 
    & Combined & \textbf{60.4} & \textbf{77.2} & \textbf{90.0} & \textbf{96.1} \\
    \cline{2-6}
    & CuBERT & 39.7 & 59.8 & 80.0 & 90.2 \\
    & CodeBERT & 36.3 & 57.1 & 78.3 & 88.8 \\
    & RoBERTa & 34.7 & 57.8 & 80.3 & 90.5 \\\midrule
\multicolumn{2}{l}{Diff (Model - Baseline)} & -20.7 & -17.4 & -10.0 &  -5.9 \\ \bottomrule
\end{tabular}}
\caption{Top-k scores for code syntax understanding. For each language, the upper block contains the results of baselines, including: (1)~\emph{Offset}: always picking the token with a fixed positional offset; (2) ~\emph{Keyword}: matching a fixed keyword nearby; and (3)~\emph{Combined}: combining the best option from \emph{Offset} and \emph{Keyword}. Score differences are calculated as the best attention score - best baseline score for each language, where a positive value indicates that the language model surpasses the baseline.}
\label{tab:main-res-PL}
\vspace{-1em}
\end{table}

\begin{table}[htb]
\scalebox{0.75}{
\begin{tabular}{@{}llllll@{}}
\toprule
\multirow{2}{*}{Language} & \multirow{2}{*}{Model} & \multicolumn{4}{l}{Top-k Score} \\
                  &                   &   k=1  &    k=3 & k=10    & k=20\\ \midrule
\multirow{3}{*}{English} 
    & Offset & 47.1 & 72.7 & 91.0 & 96.6 \\
    \cline{2-6}
    & BERT-large & \textbf{64.6} & 83.2 & 96.3 & 99.3  \\
    & RoBERTa-base & 62.7 & \textbf{84.3} & \textbf{96.9} & \textbf{99.4} \\
    & CodeBERT & 59.3 & 79.7 & 95.2 & 99.1 \\
    \midrule
\multicolumn{2}{l}{Diff (Model - Baseline)} &17.5 &  11.6 & 5.9 & 2.8   
\\ \midrule
\multirow{3}{*}{German} 
    & Offset & 36.3 & 58.0 & 83.1 & 95.1 \\
    \cline{2-6}
    & Multilingual BERT & 62.6 & 81.9 & 96.5 & 99.6  \\
    & XLM-RoBERTa-base & \textbf{67.4} & \textbf{85.5} & \textbf{97.1} & \textbf{99.7} 
\\ \midrule
\multicolumn{2}{l}{Diff (Model - Baseline)} & 31.1 & 27.5 & 14.0 & 4.6  \\ \bottomrule
\end{tabular}}
\caption{Top-k scores for NL syntax understanding. Note that BERT-large and CuBERT share the same model configuration, and CodeBERT and RoBERTa-base have the same model architecture. Unlike Table~\ref{tab:main-res-PL}, we exclude \emph{Keyword} and \emph{Combined} baselines because they do not add upon the \emph{Offset} baseline in terms of the performance.}
\label{tab:main-res-NL}
\end{table}

\begin{table}[htb]
\scalebox{0.8}{
\begin{tabular}{@{}llllll@{}}
\toprule
\multirow{2}{*}{Language} & \multirow{2}{*}{Model} & \multicolumn{4}{l}{Top-k Score (Any-token Metric)} \\
                  &                   &     k=1  &    k=3 & k=10    & k=20    \\ \midrule
\multirow{5}{*}{Python} 
    & Offset & 63.6 & 85.4 & 96.7 & 98.9 \\
    & Keyword & 22.2 & 31.3 & 34.9 & 35.2 \\ 
    & Combined & \textbf{66.8} & \textbf{88.4} & \textbf{98.2} & \textbf{99.6} \\
    \cline{2-6}
    & CuBERT & 64.3 & 82.7 & 96.1 & 99.2 \\
    & CodeBERT & 56.0 & 76.5 & 93.5 & 97.9 \\
    & RoBERTa & 49.4 & 74.7 & 94.4 & 98.5 \\\midrule
\multicolumn{2}{l}{Diff (Model - Baseline)} & -2.5 & -5.7 & -2.1 &-0.4  
\\ \midrule
\multirow{5}{*}{Java} 
   & Offset & 69.4 & 86.5 & 96.8 & 99.0   \\
    & Keyword & 40.9 & 44.9 & 46.7 & 47.0 \\
    & Combined & \textbf{75.7} & \textbf{90.0} & \textbf{98.2} & \textbf{99.6}    \\
    \cline{2-6}
    & CuBERT & 72.1 & 87.4 & 97.5 & 99.5   \\
    & CodeBERT & 62.7 & 81.1 & 93.9 & 97.6 \\
    & RoBERTa & 59.8 & 81.4 & 94.9 & 98.4 \\\midrule
\multicolumn{2}{l}{Diff (Model - Baseline)} & -3.6 & -2.6 & -0.7 & -0.1   \\ \bottomrule
\end{tabular}}
\caption{Top-k scores for code syntax understanding using the any-token metric.}
\label{tab:any-token-PL}
\vspace{-1em}
\end{table}

We present our main results to compare the performance in syntactic relation understanding on PL and NL in Tables~\ref{tab:main-res-PL} and \ref{tab:main-res-NL}, respectively. First, on \ours{}, language models generally perform worse than simple offset baseline and its combination with the keyword baseline, which indicates that the attention heads of PL pre-trained models do not effectively capture the syntactic relations in programs. The comparison between CodeBERT and RoBERTa further shows that pre-training on a large-scale code corpus, in addition to the text corpus for RoBERTa pre-training, does not yield a notably better understanding of code syntax. In comparison, language models substantially outperform offset baselines in recognizing the dependency relations in NL, demonstrating that the attention heads learn to be specialized for different relation types via large-scale pre-training on text.

Meanwhile, we present the any-token results on \ours{} in Table~\ref{tab:any-token-PL}. Although the best combined baseline still outperforms language models, the performance gap shrinks drastically. In particular, CuBERT achieves better scores than the offset baseline, and the improvement on Java is more notable. We defer the full results of different top-k scores on both PL and NL benchmarks to Appendix~\ref{app:topk}. In the following sections, we discuss the key factors that affect prediction performance.
\subsection{Case Studies: The Effect of Keywords}
\label{sec:keyword-effect}

\begin{figure}[htbp]
\centering
\vspace{-1em}
\subfigure[With semicolons (default).]{%
\includegraphics[scale=0.5]{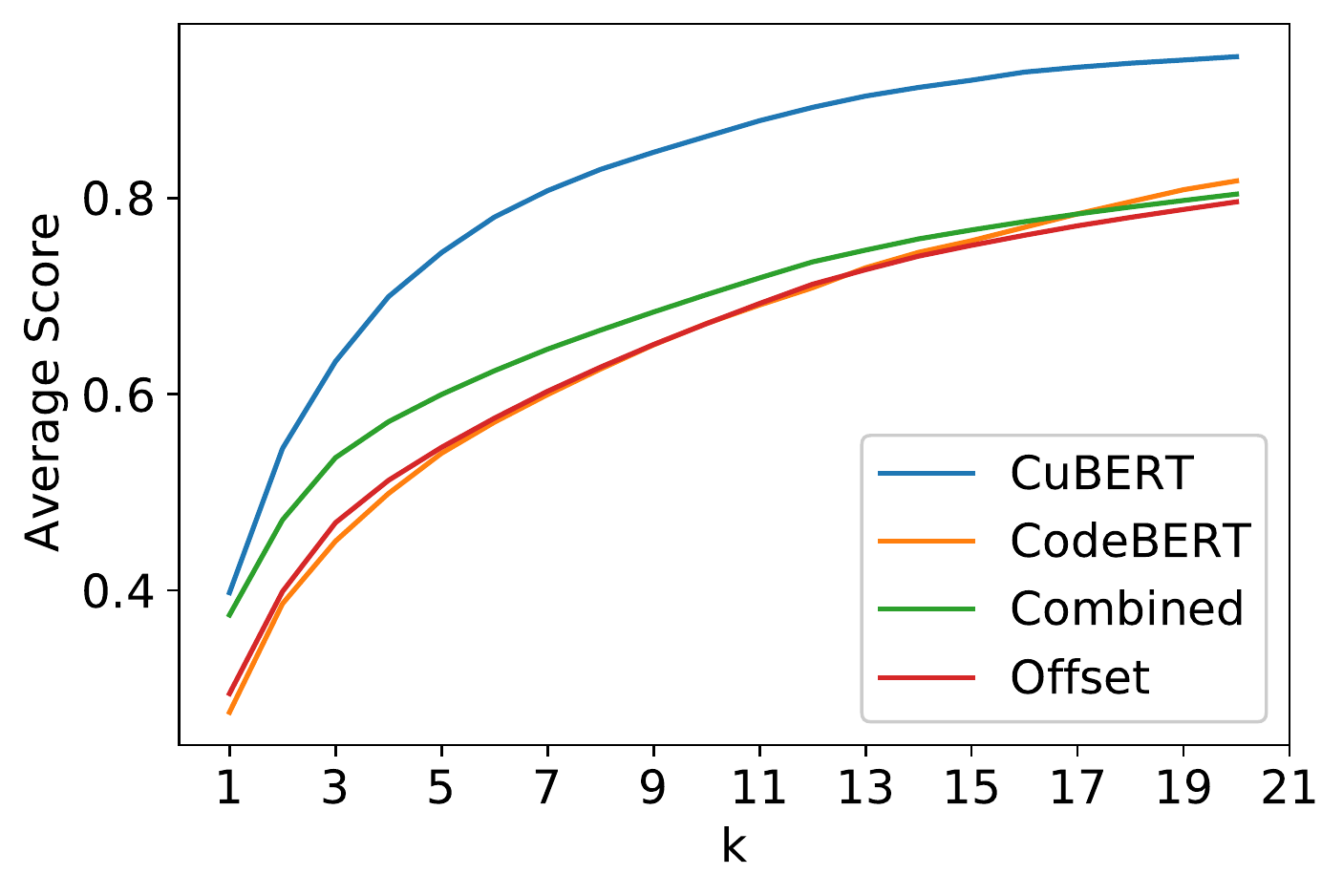}}
\vspace{-1em}
\subfigure[Without semicolons.]{%
\includegraphics[scale=0.5]{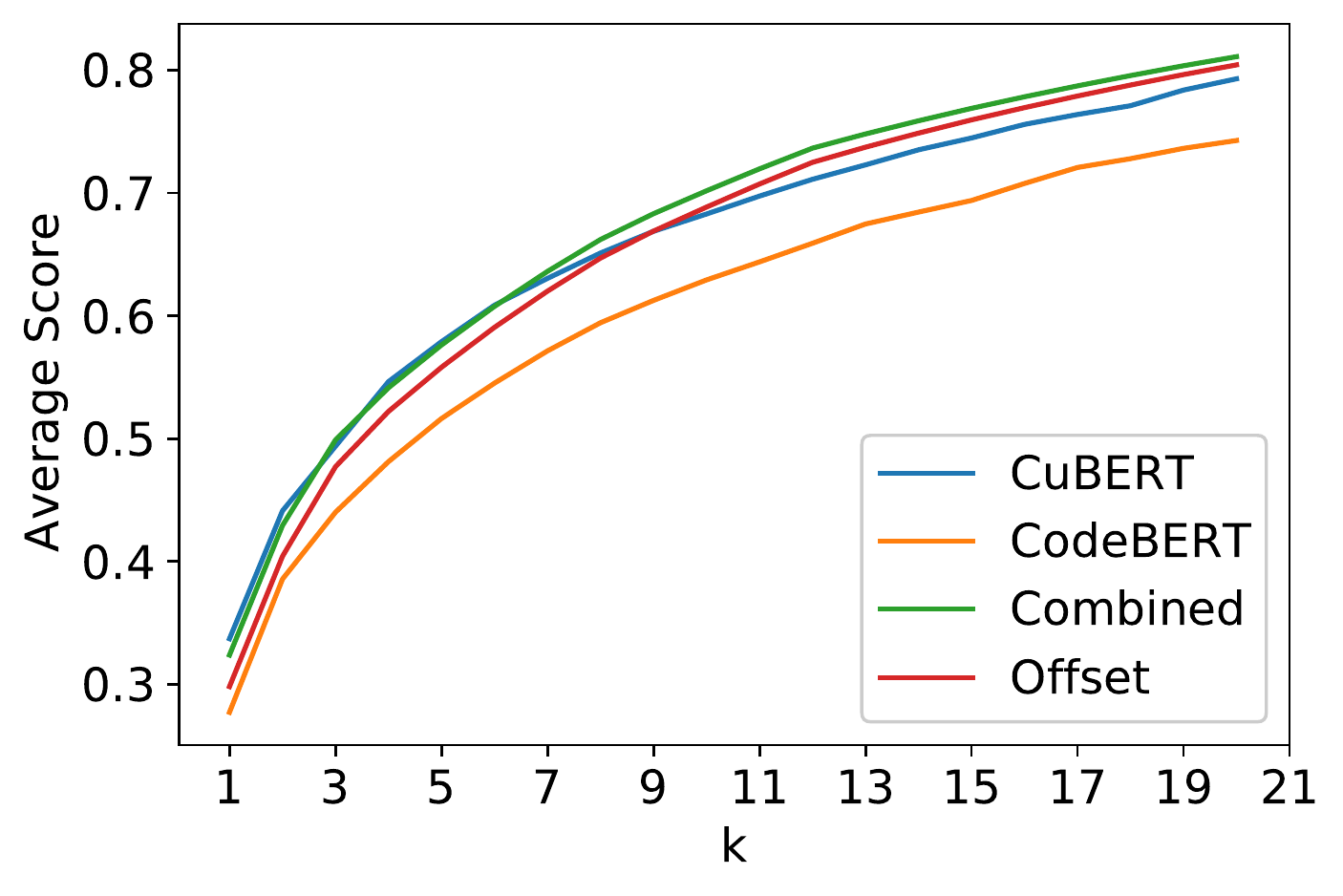}}
\caption{Top-k scores for Java syntax understanding using the last-token metric.}
\vspace{-1em}
\label{fig:ablation_java_last}
\end{figure}

To examine why the offset baseline outperforms CodeBERT and CuBERT, and why the relative performance differences get smaller when using the any-token metric, we conducted case studies and error analysis in Section \ref{sec:keyword-effect} and Section \ref{sec:error_analysis}, which both quantitatively and qualitatively categorize the error patterns. 

Firstly, we investigate the most frequently attended code tokens, and we observe that the attention heads tend to recognize the reserved tokens and keywords in PL. For example, CuBERT and CodeBERT get an improved score on Java because the semicolon token is part of the ground truth dependent node, which is a popular token attended to by language models. Based on this observation, we perform an ablation study on the presence of the semicolon in ground truth annotations. When the semicolon tokens are removed from ground truth dependent nodes, we also disable the language models to attend to semicolons in the input code. Since the semicolon appears at the end of each Java statement, here we compute the last-token score which may be significantly affected by semicolons. As shown in Figure~\ref{fig:ablation_java_last}, CuBERT substantially outperforms baselines when semicolons are included in the ground truth labels. On the other hand, CuBERT reaches lower scores than baselines when semicolons are excluded from ground truth labels and predictions. The comparison suggests that attention heads are more capable of identifying frequent keywords in the model input. We defer the full ablation study on both Python and Java to Appendix~\ref{app:separation-tokens}.

We further discuss the breakdown results with respect to relation types, and we select some representative relations for Python that highlight the performance differences between CuBERT and the offset baseline in Table~\ref{tab:attn_offset_python_first_selected}. First, the attention is highly capable of performing keyword matching, which leads to decent accuracy on relations that connect popular keywords, such as \texttt{If:if$\rightarrow$else}. However, when the head and dependent tokens are diverse, it becomes challenging for the language model to recognize the relation. For example, in relation types \texttt{Assign:target$\rightarrow$value} and \texttt{Call:func$\rightarrow$args}, both head and dependent nodes can take various identifier names defined by different programmers. In particular, CuBERT can not effectively utilize the relative positions of tokens to learn the relations, even if the dependent node is near the head node. In such situations, the offset baseline with a fixed offset value of 2 already surpasses the pre-trained model. The full breakdown results of all relation types on both Python and Java can be found in Appendix~\ref{app:relation-breakdown}.

\begin{table}[htbp]
\setlength\tabcolsep{2pt}
\scalebox{0.95}{
\begin{tabular}{@{}lllll@{}}
\toprule
 \multirow{2}{*}{Relation} & \multicolumn{2}{c}{Score} & \multirow{2}{*}{Offset} & \multirow{2}{*}{Diff} \\
     &     CuBERT      &     Offset     &  &      \\ \midrule
If:if$\rightarrow$else & 92.7 & 5.7 & 17 & 87.1 \\
If:body$\rightarrow$orelse & 29.2 & 7.1 & 12 & 22.0 \\
If:if$\rightarrow$body & 31.5 & 23.1 & 7 & 8.4 \\
For:for$\rightarrow$body & 30.4 & 32.7 & 7 & -2.3 \\
Assign:target$\rightarrow$value & 39.8 & 71.2 & 2 & -31.4 \\
While:test$\rightarrow$body & 16.2 & 48.5 & 4 & -32.4 \\
Call:func$\rightarrow$args & 59.3 & 93.2 & 2 & -33.9 \\
 \bottomrule
\end{tabular}}
\caption{The comparison of top-1 first-token scores between CuBERT and the offset baseline with the best fixed offset for selected relation types on Python dataset.}
\label{tab:attn_offset_python_first_selected}
\vspace{-1em}
\end{table}

\subsection{Error Analysis}
\label{sec:error_analysis}

\begin{table}[htb]
\vspace{-1em}
\scalebox{0.8}{
\setlength\tabcolsep{2pt}
\begin{tabular}{@{}lp{5cm}cc@{}}
\toprule
\multirow{2}{*}{Relation} & \multicolumn{1}{c}{\multirow{2}{*}{Error Situation}} & \multicolumn{2}{c}{Count} \\
 & \multicolumn{1}{c}{}   & Python       & Java       \\ \midrule
\multirow{3}{*}{\makecell{If:\\if$\rightarrow$else}}  &   Nested if statements or multiple if statements close to each other.
  & 34   &  42  \\\cline{2-4}
& Predicts other keywords inside body block, e.g., \texttt{if} and \texttt{while}.  & 11 & 4 \\\cline{2-4}
& Other. & 5& 4 \\
\midrule
\multirow{4}{*}{\makecell{If:\\body$\rightarrow$orelse}}  &  Predicts another token with the same name as head token itself.   &  38  & 21  \\\cline{2-4}
& Predicts keywords inside body block, e.g., \texttt{if}, \texttt{;} and \texttt{while}. & 0 & 14 \\\cline{2-4}
& Predicts a long string or docstring. & 7 & 7 \\\cline{2-4}
& Other. & 5 & 8 \\
\midrule
\multirow{3}{*}{\makecell{If:\\if$\rightarrow$body}}  &   Predicts blank space or tab.  &  32  & 0  \\\cline{2-4}
& Predicts \texttt{\{} or \texttt{\}}. & 0 & 27 \\\cline{2-4}
& Predicts \texttt{return}. & 0 & 19 \\\cline{2-4}
& Predicts \texttt{\textbackslash n} or \texttt{:}. & 15 & 0 \\\cline{2-4}
& Other. & 3 & 4 \\
\midrule
\multirow{3}{*}{\makecell{For:\\for$\rightarrow$body}}  &  Predicts blank space or tab.    &   46 & 0  \\\cline{2-4}
& Predicts \texttt{\{} or \texttt{\}}.  & 0 & 29 \\\cline{2-4}
& Other. & 4 & 21 \\
\midrule
\multirow{3}{*}{\makecell{Assign:\\target$\rightarrow$value}}  &  Predicts a token that comes before \texttt{=}, e.g. \texttt{a[0]} and \texttt{a.b}    &   16 &  30  \\\cline{2-4}
& Predicts \texttt{=}. & 22 & 5 \\\cline{2-4}
& Other. & 12 & 15 \\
\midrule
\multirow{2}{*}{\makecell{While:\\test$\rightarrow$body}}  &  Predicts a token in the test block.   &  48  & 36  \\\cline{2-4}
& Other. & 2 & 14 \\
\midrule
\multirow{3}{*}{\makecell{Call:\\func$\rightarrow$args}}  &  Predicts \texttt{(} or \texttt{)}.    & 45   & 37  \\\cline{2-4}
& Predicts another token with the same name as head token itself. & 0 & 10 \\\cline{2-4}
& Other. & 5& 3 \\
\bottomrule
\end{tabular}
}
\caption{Error analysis using CuBERT.}
\label{tab:error-analysis}
\end{table}

To categorize the wrong predictions of the attention, we manually examine 50 error cases for each relation selected in Table~\ref{tab:attn_offset_python_first_selected}, and present the error situations in Table~\ref{tab:error-analysis}. Again, we observe that the attention often incorrectly selects frequently occurring tokens such as brackets. Moreover, the model has difficulty capturing the hierarchical code structure, thus it often attends to nearby keywords regardless of logical code blocks.

% \begin{verbatim}
% if e.errno == errno.EEXIST:
%     pass
% else:
%     ...
% \end{verbatim}

% \begin{verbatim}
% if (len(self.stack)>0): 
%     while ...
%         ...
% else: 
%     ...
% \end{verbatim}

% \begin{figure}[htbp]
% \centering
% \vspace{-1em}
% \includegraphics[width=0.9\linewidth]{figures/code_if_else.PNG}
% \vspace{-1em}
% \caption{If: if $\rightarrow$ else.}\label{code_if_else}
% \vspace{-1em}
% \end{figure}

% \begin{figure}[htbp]
% \centering
% \vspace{-1em}
% \includegraphics[scale=0.5]{figures/Python Head 17-2 if else.pdf}
% \vspace{-1em}
% \caption{Python Head 17-2 If: if $\rightarrow$ else.}\label{case_if_else}
% \vspace{-1em}
% \end{figure}

\begin{figure}[ht]
\centering
\vspace{-1em}
\subfigure[Python code.]{%
\includegraphics[scale=0.55]{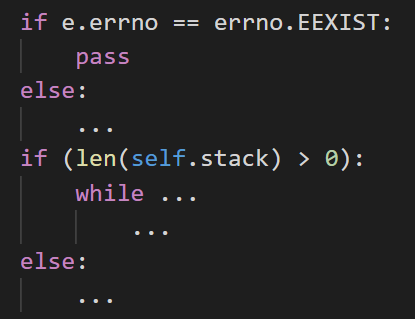}}
\vspace{-1em}
\subfigure[Attention weights.]{%
\includegraphics[scale=0.5]{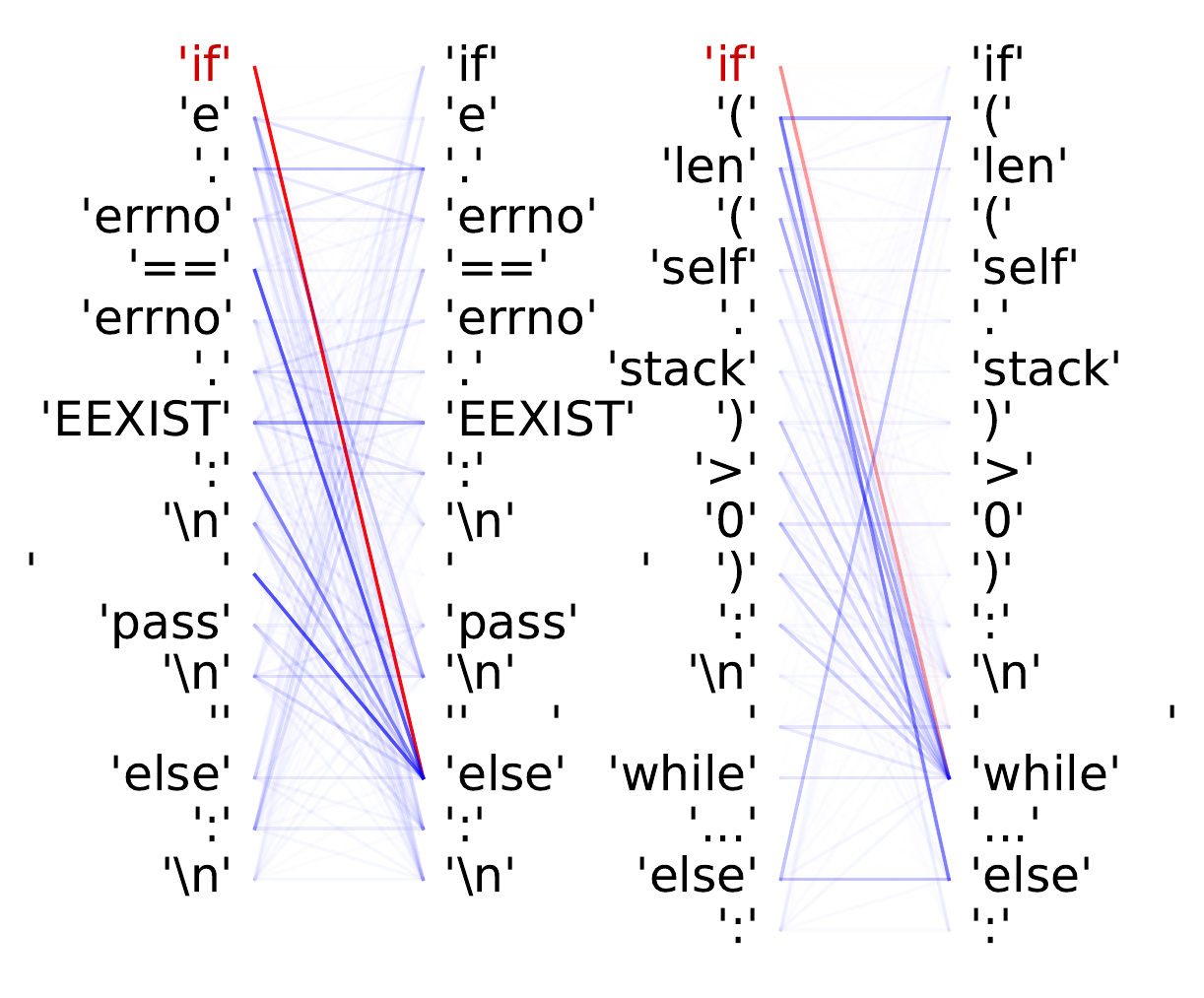}}
\caption{Two sample cases for the relation \texttt{If:if$\rightarrow$else} and corresponding attention weights of CuBERT's head 17-2.}
\label{case_if_else}
\vspace{-1em}
\end{figure}

Take the relation \texttt{If:if$\rightarrow$else} as an example, on which the language model generally achieves the best performance. Shown in Figure~\ref{case_if_else} are two sample \texttt{if}-statements, where the first one does not contain nested flow control blocks while the second one contains a keyword \texttt{while} inside the \texttt{if}-body. "..." denotes that some code is omitted. Visualizing their corresponding attention weights of the attention head that performs the best on the relation \texttt{If:if$\rightarrow$else}, we observe that the attention head correctly attends to the \texttt{else} token in the first example, while it wrongly attends to the \texttt{while} token inside the \texttt{if}-body in the second example. More examples like these can be found in Appendix~\ref{app:predictions}.

\section{Related Work}

% \paragraph{Language models for natural language processing.}

% One approach is to investigate the outputs of language models on carefully chosen input sentences. Some other authors analyze the internal representations of models.

Transformer-based language models have been widely used for natural language processing~\cite{devlin2019bert, DBLP:journals/corr/abs-1907-11692,wang2020language,wang2021zero,shen-etal-2022-lass,wang-etal-2022-deepstruct}. \citet{DBLP:conf/naacl/HewittM19} show that syntax trees are implicitly embedded in BERT's word representation space via a structural probe. Another line of work studies what is learned by the attention in language models~\cite{clark2019does,raganato2018analysis,DBLP:conf/acl/VoitaTMST19,DBLP:conf/nips/MichelLN19, DBLP:journals/corr/abs-1904-02679, DBLP:conf/emnlp/BurnsNGGG18, DBLP:conf/emnlp/MarecekR18, DBLP:conf/acl/TitovSSV18}. In particular,~\citet{clark2019does} evaluate the attention heads of BERT on dependency parsing tasks using the English Penn Treebank corpus, where the attention significantly outperforms offset baselines. 
On the contrary, we demonstrate that attention-based models largely perform worse than offset baselines on code syntax understanding.

% \paragraph{Language models for programming languages.}
The success of Transformer-based models for natural language processing leads to their application in the PL domain~\cite{kanade2020learning,feng2020codebert, DBLP:conf/nips/RoziereLCL20, DBLP:journals/corr/abs-2110-06773, DBLP:conf/emnlp/ClementDTSS20, DBLP:conf/iclr/DehghaniGVUK19}.
\citet{DBLP:journals/corr/abs-2107-03374} evaluate the model performance by measuring the functional correctness on unit tests. \citet{10.1145/3468264.3468611} empirically shows that Transformers can utilize syntactic information to make predictions in some code processing tasks, while we analyze attention's ability to understand syntactic relations. 
\citet{DBLP:conf/kbse/KarmakarR21} probe pre-trained models on four code understanding tasks. They focus more on code classification, e.g., they train a classifier for predicting the AST node tag and the code length. On the contrary, we probe the attention heads for syntactic relation understanding, and we aim to present a comprehensive study of the differences between pre-trained language models on NL and PL for capturing the syntax structures.
% We believe our benchmark benefits future studies of whether we are able to train models for these specific tasks.

There have been some efforts that try to take code structure into account during pre-training of Transformer-based models for code. For example, GraphCodeBERT~\cite{DBLP:conf/iclr/GuoRLFT0ZDSFTDC21} utilizes data flow for pretraining; i.e., the relation of "where-the-value-comes-from" for variables. On our Python benchmark, GraphCodeBERT achieves a top-1 first-token score of 39.3, which is better than 33.1 of CodeBERT, and comparable to 39.2 of CuBERT. However, such a score is still worse than 43.6 of the offset baseline. This trend is consistent when evaluating with other metrics.  These results show that pre-training on data flow helps improve the model's ability to understand code syntax, but there is still large room for improvement.

\section{Conclusion}

In this work, we introduce \ours{}, a large-scale benchmark for measuring the performance of code syntax understanding. Based on \ours{}, we conduct the first comprehensive study to analyze the capability of pre-trained language models on understanding the code syntactic structures without further finetuning. We demonstrate that while the attention heads of pre-trained language models are able to identify dependency relations in NL to some extent, they have difficulty recognizing the syntactic relations in programs. Pre-trained models even generally perform worse than simple offset baselines, and they tend to attend to frequently occurring nearby tokens without taking the hierarchical code structure into consideration.

We also analyze the differences between NL and PL from the perspectives of pre-trained models. Our evaluation suggests that PL has unique characteristics that distinguish them from NL, such as the long-term dependency between code tokens, and the hierarchy in the syntactic structures. Therefore, simply taking a program as a token sequence is insufficient for modeling the program structure, which could eventually limit the potential of language models for code understanding tasks. We consider developing new model architectures and pre-training algorithms to leverage and represent the code structure and dependency graph as important future work.

\section{Limitations}
% % Although we use three different metrics and discover a relatively reliable tendency across all the metrics, the metrics themselves are not perfect and can be improved. 
% We only experiment with six models for two natural languages and two programming languages. More experiments are needed to extend the results to other pre-trained models and languages in the PL domain. 
% Because we provide an automated workflow for extracting syntactic relations,
% it is not difficult to construct datasets for other programming languages when there are AST parsers available.
% % Though we identify pre-trained language models' inability to understand code syntax, we do not have a complete solution to it. 
% We call for community collaboration to further investigate and resolve this issue.

For the limitations of our benchmark, the gold annotations are based on the AST parsers. Adding new programming languages whose parsers are unavailable will require additional labeling efforts. A limitation in our experimental setup is that we have only benchmarked six models across two kinds of natural languages and programming languages. Finally, the main focus of our study is to probe the language models for code understanding. As a result, we have not proposed models that could deal with the code syntax in natural language and programming language applications. Future work could include developing such models that capture both semantics and structures.

\section{Ethical Considerations}
We hereby acknowledge that all of the co-authors of this work are aware of the provided \textit{ACM Code of Ethics} and honor the code of conduct. The followings give the aspects of both our ethical considerations and our potential impacts to the community. This work creates a benchmark to test the code syntax understanding of pre-trained language models. Instead of natural language, the programming language is used for pre-training. We do not anticipate the production of harmful outputs after using our benchmark and existing models, especially towards vulnerable populations. 

\section{Environmental Considerations}
We use several pre-trained language models. According to the estimation in \cite{strubell-etal-2019-energy}, pre-training a model with a similar size as used in the work costs 1,507 kWh$\cdot$PUE and emits 1,438 lb $CO_2$. This work focuses on inference. Therefore, our energy cost and $CO_2$ emissions are relatively small.

\section*{Acknowledgements}
We would like to thank the anonymous reviewers for their suggestions and comments. This material is in part based upon work supported by Berkeley DeepDrive and Berkeley Artificial Intelligence Research.

\bibliography{custom}
\bibliographystyle{acl_natbib}

% \newpage
\appendix

\begin{figure*}[htbp]
\centering
\subfigure[Python (First Token Metric)]{%
\includegraphics[width=0.45\linewidth]{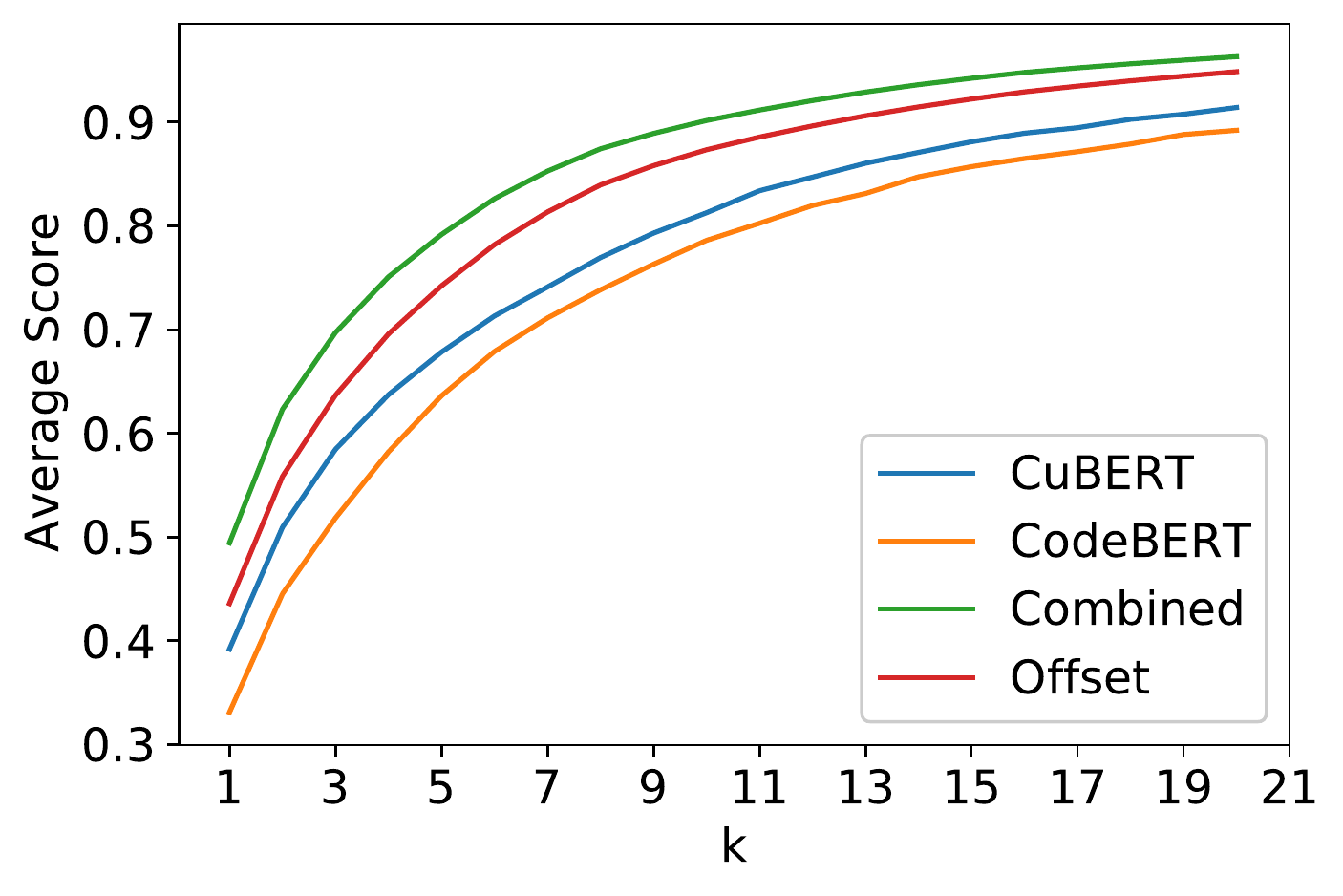}
\label{fig:python_test_first}} 
\hspace*{-3mm}
\subfigure[Java (First Token Metric)]{%
\includegraphics[width=0.45\linewidth]{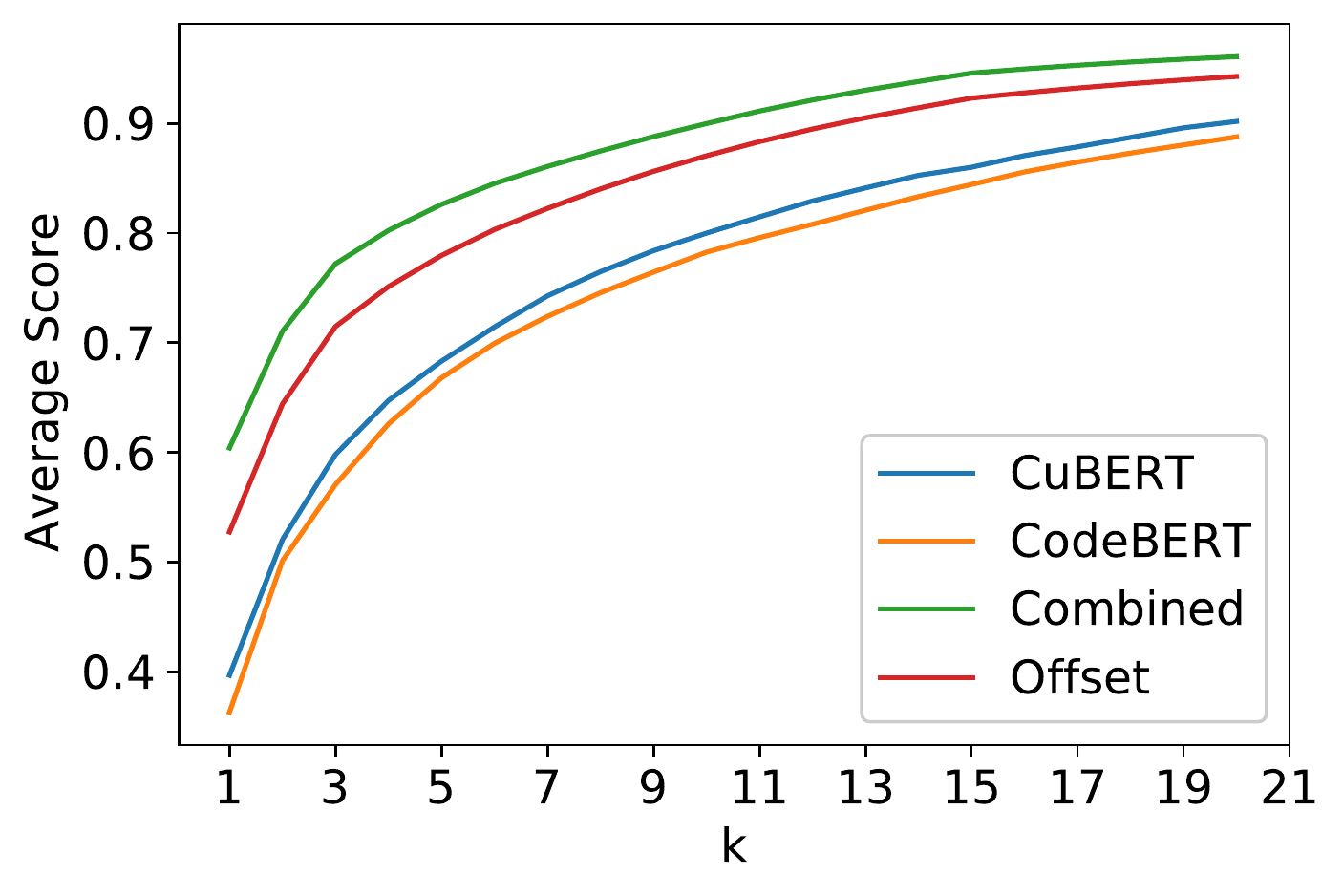}
\label{fig:java_test_first}}

\subfigure[Python (Any Token Metric)]{%
\includegraphics[width=0.45\linewidth]{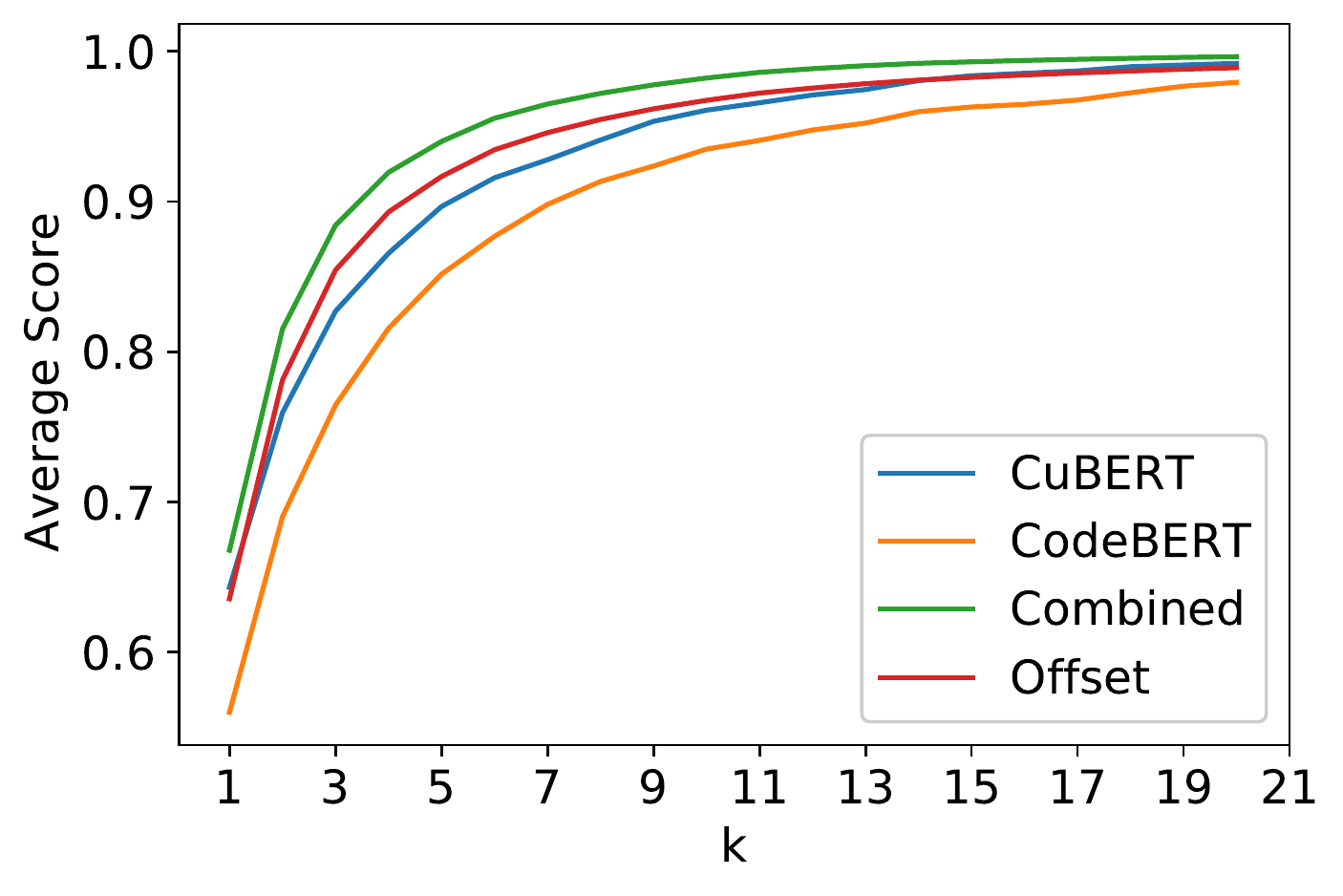}
\label{fig:python_test_any}} 
\hspace*{-3mm}
\subfigure[Java (Any Token Metric)]{%
\includegraphics[width=0.45\linewidth]{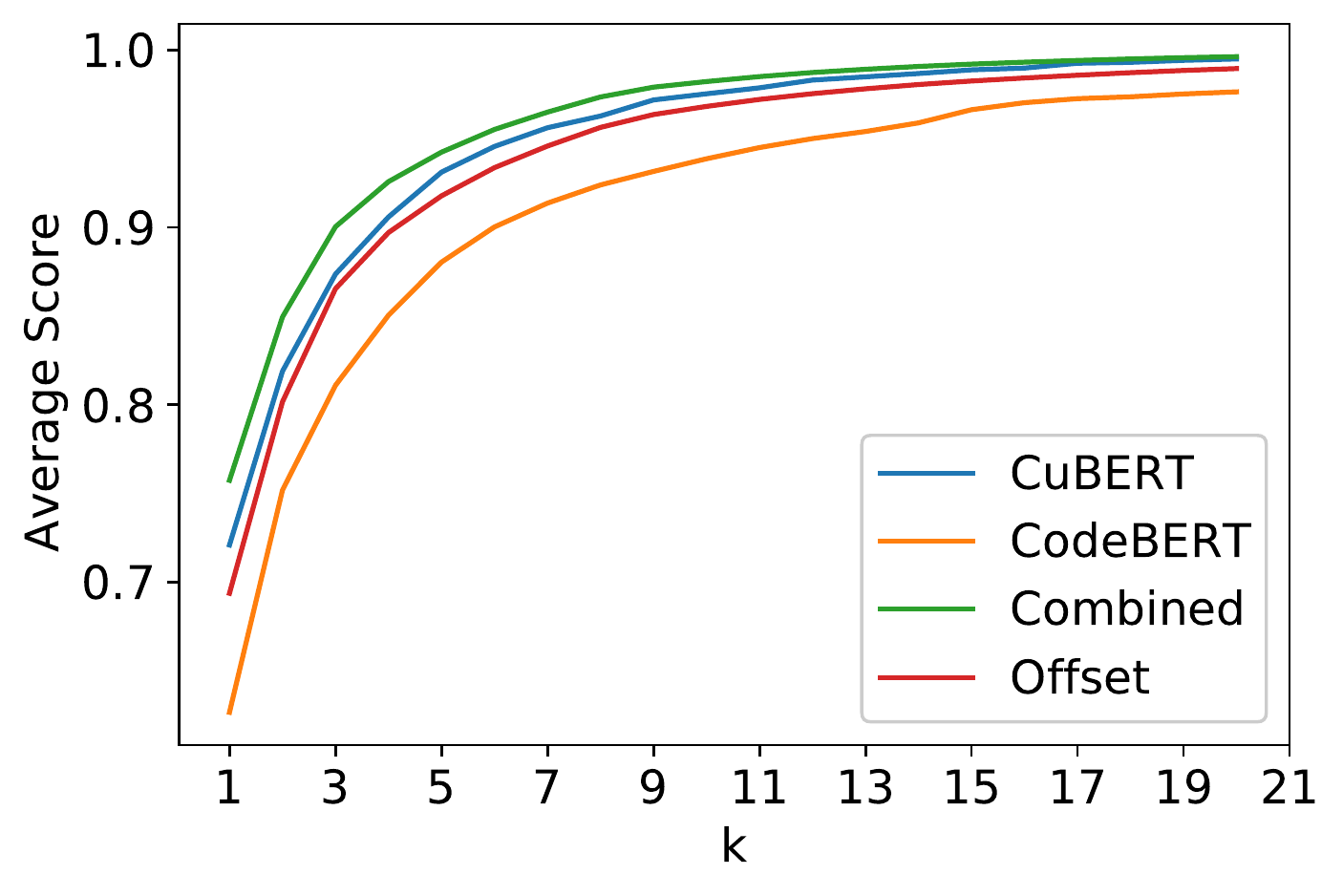}
\label{fig:java_test_any}}

\caption{PL Top-k Scores On Test Set} \label{fig:PL_topk}
\end{figure*}

\begin{figure}[htbp]
\centering
\includegraphics[width=\linewidth]{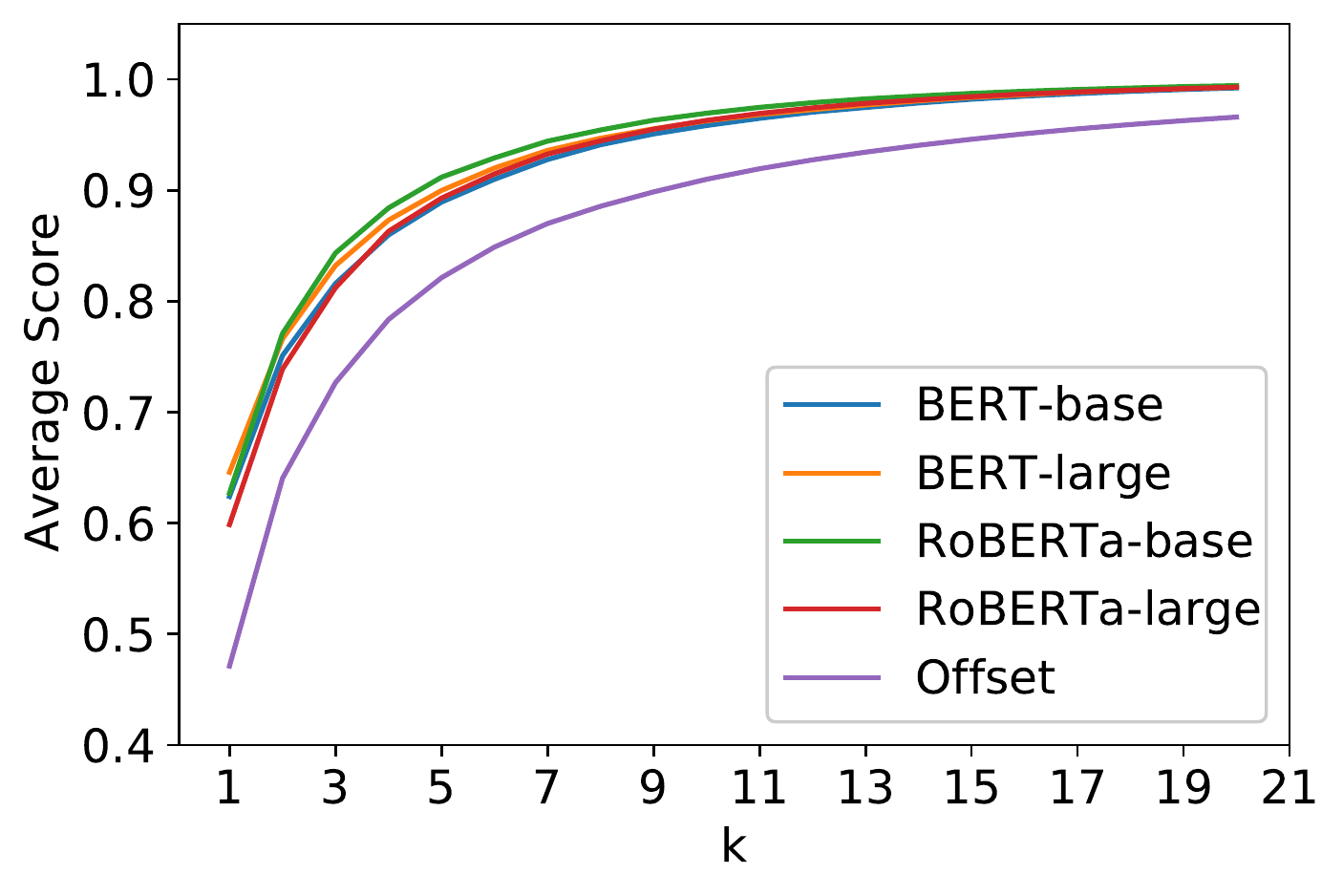}
\caption{English Top-k Scores}
\label{fig:english}
\vspace{-1em}
\end{figure}

\begin{figure}[htbp]
\centering
\includegraphics[width=\linewidth]{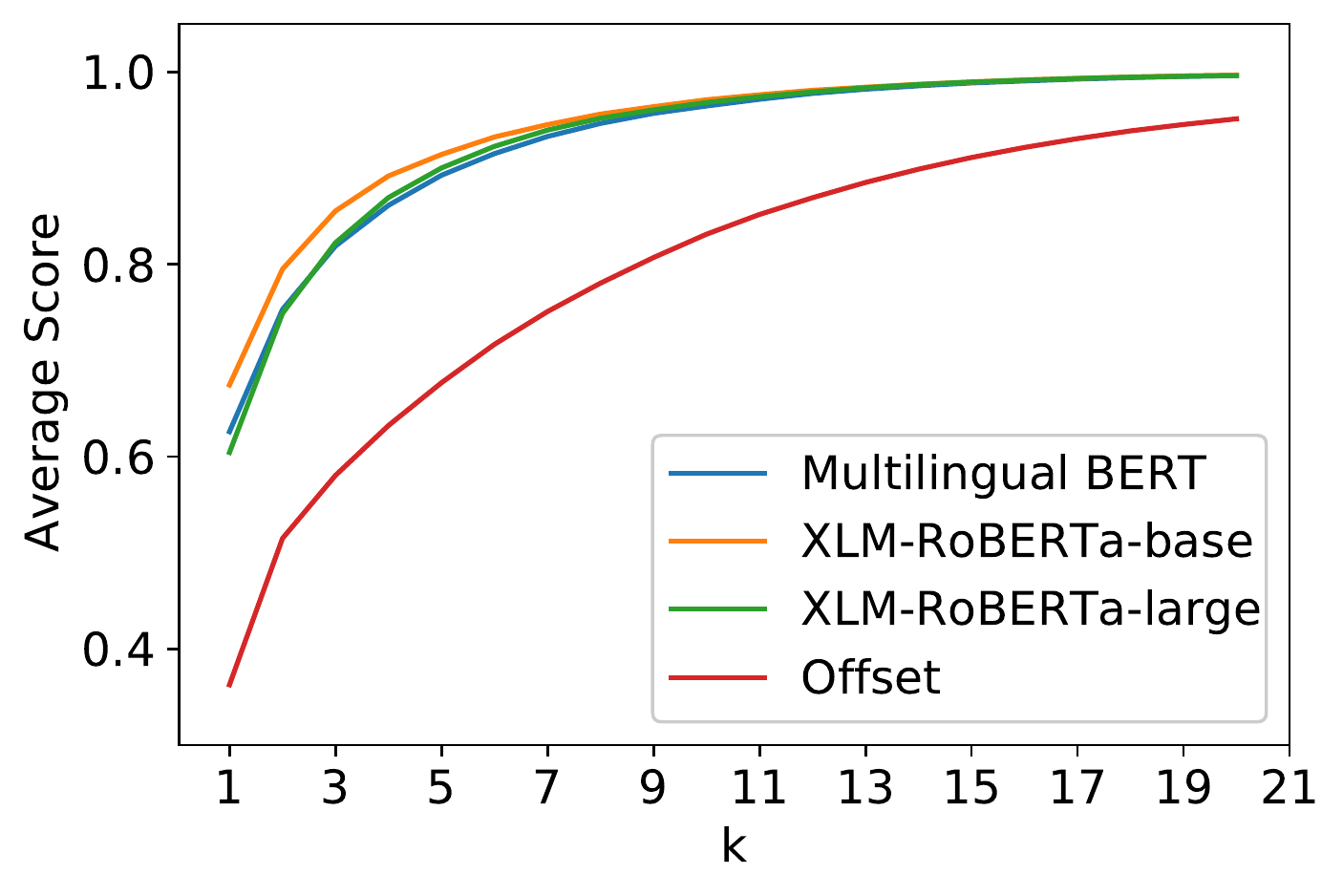}
\caption{German Top-k Scores}
\label{fig:german}
\vspace{-1em}
\end{figure}

\section{More Details on \ours{} Construction}
\label{app:data-construction}
Since the code search net dataset does not come with syntactic relation labels, we come up with a way of extracting syntactic relations. We first utilize python's tokenize module and javalang module to produce code tokens from source code, and then label these code tokens with syntactic relations by using AST parsers on source code. We utilize Python ast module~\cite{pythonast2021} and Java org.eclipse.jdt.core.dom.ASTParser class~\cite{javaast2014} to parse source code into ast nodes. The AST structure captures syntactical relations. An AST node has children AST nodes and a name that denotes its class.  We use the class of the node as label and children nodes as dependents and heads when generating annotations. For example, the source code \texttt{A = B}, which means assigning value \texttt{B} to target variable \texttt{A}, is parsed into the AST node \texttt{Assign(targets=[Name(id='A')], value=Name(id='B'))}. It gives us a syntactic relation whose head is \texttt{A} and dependent is \texttt{B}, annotated with the relation type label \texttt{Assign}. Full statistics of \ours{} are displayed in Table~\ref{tab:data_stats_full}.

\begin{table*}[ht]
\small
\setlength\tabcolsep{3pt}
\scalebox{0.95}{
\begin{tabular}{lrrp{4cm}ll}
\toprule
\multirow{2}{*}{Relation:head$\rightarrow$dependent} & \multicolumn{2}{c}{Count} & \multirow{2}{*}{Explanation}         & \multicolumn{2}{l}{~~~~~Code Example}  \\
  & Python      & Java        &      & Python           & Java            \\
 \midrule
Assign:target$\rightarrow$value & 78482 & 13384 & Assigning a value to a target variable. & \textbf{\color{blue}target} = \textbf{\color{red}10} & int \textbf{\color{blue}target} = \textbf{\color{red}10}; \\
\hline
\multirow{3}{*}{Attribute:value$\rightarrow$attr} & \multirow{3}{*}{158797} & \multirow{3}{*}{84215} & Accessing the attribute (member field or member function) of an value. & \multirow{3}{*}{\boldblue{value}.\boldred{attribute}} & \multirow{3}{*}{\boldblue{line}.\boldred{setLength}(2);}\\
\hline
\multirow{4}{*}{AugAssign:target$\rightarrow$value} & \multirow{4}{*}{3150} & \multirow{4}{*}{/} & An assignment augmented with an operation. For Java, this case is included in Assign:target$\rightarrow$value. & \multirow{4}{*}{\boldblue{x} += \boldred{2}} & \multirow{4}{*}{/} \\
\hline
BinOp:left$\rightarrow$right & 26035 & / & A binary operation. & \boldblue{a} + \boldred{b} & / \\
\hline
BoolOp:value$\rightarrow$value & 5783 & / & A boolean operation & \boldblue{True} or \boldred{False} & / \\
\hline
Call:args$\rightarrow$keywords & 9256 & / & \multirow{3}{4cm}{Calling a function with some arguments (and keywords).} & \multirow{3}{*}{\textbf{\color{blue}function}(\boldgreen{arg}, \textbf{\color{red}key=1})} & \multirow{3}{*}{\textbf{\color{blue}function}(\textbf{\color{red}arg});} \\
Call:func$\rightarrow$args & 110949 & 50890 & & & \\
Call:func$\rightarrow$keywords & 16274 & / & & & \\
\hline
Compare:left$\rightarrow$comparator & 25852 & / & A comparison between values. & \boldblue{a} < \boldred{b} & / \\
\hline
Dict:key$\rightarrow$value & 7787 & / & Initializing a dictionary. & \{\boldblue{count} : \boldred{10} \}& / \\
\hline
DictComp:key$\rightarrow$generator & 359 & / &  & & \\
DictComp:key$\rightarrow$value & 359 & / & Dictionary comprehension. & \{\boldblue{i}: \boldgreen{2*i} for \boldred{ i in list}\} & /\\
DictComp:value$\rightarrow$generator & 359 & / & & & \\
\hline
Do:body$\rightarrow$test & / & 38 & \multirow{3}{4cm}{The do loop repeatedly executes the body block as long as the condition is true.} & & \multirow{3}{*}{\makecell{\boldblue{do} \\ ~~~~~\boldgreen{statement;} \\while (\boldred{condition});}} \\
Do:do$\rightarrow$body & / & 45 & & /& \\
Do:do$\rightarrow$test & / & 38 & & & \\
\hline
For:for$\rightarrow$body & 8704 & 1864 & \multirow{15}{4cm}{A for loop repeatedly executes the body block for some iterations.} & \multirow{15}{*}{\makecell{\textbf{\color{blue}for} \boldgreen{target} in \boldgreen{iter}:\\ ~~~~~\textbf{\color{red}body}}}& \multirow{15}{*}{\makecell{\textbf{\color{blue}for} (\boldgreen{initializers}; \\~~~~~\boldgreen{test}; \boldgreen{updaters}) \{\\ ~~~~~\textbf{\color{red}body;} \\ \}}} \\
For:for$\rightarrow$initializers & / & 1650 & & & \\
For:for$\rightarrow$iter & 8704 & / & & & \\
For:for$\rightarrow$target & 8704 & / & & & \\
For:for$\rightarrow$test & / & 1296 & & & \\
For:for$\rightarrow$updaters & / & 1682 & & & \\
For:initializers$\rightarrow$body & / & 1781 & & & \\
For:initializers$\rightarrow$test & / & 1286 & & & \\
For:initializers$\rightarrow$updaters & / & 1670 & & & \\
For:iter$\rightarrow$body & 8704 & / & & & \\
For:target$\rightarrow$body & 8704 & / & & & \\
For:target$\rightarrow$iter & 8704 & / & & & \\
For:test$\rightarrow$body & / & 1789 & & & \\
For:test$\rightarrow$updaters & / & 1678 & & & \\
For:updaters$\rightarrow$body & / & 1685 & & & \\
\hline
GeneratorExp:elt$\rightarrow$generator & 685 & / & A generator expression. & (\boldblue{2*i} for \boldred{ i in list})& / \\
\hline
If:body$\rightarrow$orelse & 11024 & 4976 & \multirow{6}{4cm}{An if statement conditionally executes a body based upon some criteria.} & \multirow{6}{*}{\makecell{\textbf{\color{blue}if} \boldgreen{condition}:\\ ~~~~~\boldgreen{body1} \\ \textbf{\color{red}else}: \\ ~~~~~\boldred{body2}}}& \multirow{6}{*}{\makecell{\textbf{\color{blue}if} (\boldgreen{condition}) \{\\ ~~~~~\boldgreen{body1;} \\  \} \textbf{\color{red}else} \{ \\ ~~~~~\boldred{body2;} \\\} }} \\
If:if$\rightarrow$body & 34250 & 22392 & & & \\
If:if$\rightarrow$else & 11024 & 5038 & & & \\
If:if$\rightarrow$test & 34250 & 19323 & & & \\
If:test$\rightarrow$body & 34250 & 22392 & & & \\
If:test$\rightarrow$orelse & 11024 & 5007 & & & \\
\hline
IfExp:body$\rightarrow$orelse & 1262 & 1173 & \multirow{4}{4cm}{An if expression (conditional expression).} & \multirow{4}{*}{\boldblue{x} if \boldgreen{condition} else \boldred{y}} & \multirow{4}{*}{(\boldblue{condition}) ? \boldgreen{x} : \boldred{y}}\\
IfExp:body$\rightarrow$test & 1262 & / & & & \\
IfExp:test$\rightarrow$body & / & 1218 & & & \\
IfExp:test$\rightarrow$orelse & 1262 & 1173 & & & \\
\hline
InfixExpr:left$\rightarrow$right & / & 35170 & Infix expression of the form leftOperand InfixOperator rightOperand. & / & \boldblue{a} + \boldred{b} \\
\hline
InstanceofExpr:expr$\rightarrow$type & / & 1367 & Checking whether an expression is  some type. & / & \boldblue{input} instanceof \boldred{String}\\
\hline
LabeledStatement:label$\rightarrow$body & / & 10 & A statement labeled with an identifier. & /& \boldblue{Identifier} : \boldred{Statement} \\
\hline
ListComp:elt$\rightarrow$generator & 2691 & / & List comprehension. & [\boldblue{x} for \boldred{x in list1}] & /\\
\hline
SetComp:elt$\rightarrow$generator & 67 & / & Set comprehension. & \{\boldblue{x} for \boldred{x in list1}\} & /\\
\hline
Slice:lower$\rightarrow$upper & 731 & / & A slice used in subscript of lists. & A[\boldblue{2}:\boldred{6}] & /\\
\hline
Subscript:value$\rightarrow$slice & 39271 & 4555 & Accessing parts of an array or data structure through subscript. & \boldblue{A}[\boldred{2:6}]& \boldblue{A}[\boldred{0}] \\
\midrule
&&&&& Continued on next page.\\
\bottomrule
\end{tabular}
}
% \caption{Full dataset statistics table of \ours{}. For each relation type, we highlight the head and dependent nodes in the examples in bold, with the head in blue and the dependent in red. If a node can be either head or dependent in different relations, we color it in green. For more explanation about the syntax, please refer to the documentation of Python ast module~\cite{pythonast2021} and Java org.eclipse.jdt.core.dom.ASTParser~\cite{javaast2014}}
% \label{tab:data_stats_full}
\end{table*}

\begin{table*}[ht]
\small
\setlength\tabcolsep{3pt}
\scalebox{0.95}{
\begin{tabular}{lrrp{4cm}ll}
\toprule
\multicolumn{2}{c}{Continued from previous page.} &&&& \\
\midrule
\multirow{2}{*}{Relation:head$\rightarrow$dependent} & \multicolumn{2}{c}{Count} & \multirow{2}{*}{Explanation}         & \multicolumn{2}{l}{~~~~Code Example}  \\
  & Python      & Java        &      & Python           & Java            \\
 \midrule
Switch:expr$\rightarrow$statement & / & 385 & \multirow{3}{4cm}{A switch statement chooses a branch to execute based upon conditions.} & & \multirow{3}{*}{\makecell{\boldblue{switch} (\boldgreen{inputExpr}) \{\\     ~~~~~\boldred{Statement;}\\ \} }}\\
Switch:switch$\rightarrow$expr & / & 320 & & /& \\
Switch:switch$\rightarrow$statement & / & 385 & & & \\
\hline
Try:body$\rightarrow$finalbody & 135 & 474 & \multirow{6}{4cm}{A try statement for handling exceptions.} & \multirow{6}{*}{\makecell{try:\\~~~~~\boldblue{body1}\\except \boldgreen{Exception:}\\~~~~~\boldgreen{body2}\\else:\\~~~~~\boldred{body3}\\finally:\\~~~~~\boldred{body4}}}& \multirow{6}{*}{\makecell{try \{\\~~~~~\boldblue{body1;}\\ \} \boldgreen{catch (Exception e)} \{\\~~~~~\boldgreen{body2;}\\ \} finally \{ \\~~~~~\boldred{body3;}\\ \}}}\\
Try:body$\rightarrow$handler & 3020 & 2011 & & & \\
Try:body$\rightarrow$orelse & 181 & / & & & \\
Try:handler$\rightarrow$finalbody & 48 & 186 & & & \\
Try:handler$\rightarrow$orelse & 181 & / & & &  \\
&&&&&\\
&&&&&\\
&&&&&\\
\hline
While:test$\rightarrow$body & 743 & 975 & \multirow{4}{4cm}{The while loop repeatedly executes the body block as long as the condition is true.}& \multirow{4}{*}{\makecell{\boldblue{while} \boldgreen{condition}:\\ ~~~~~\textbf{\color{red}body}}}& \multirow{4}{*}{\makecell{\boldblue{while} (\boldgreen{condition}) \{\\ ~~~~~\textbf{\color{red}body;} \\ \}}} \\
While:while$\rightarrow$body & 743 & 975 & & & \\
While:while$\rightarrow$test & 743 & 416 & & & \\\\
\hline
With:item$\rightarrow$body & 1239 & / & A with statement with built-in context manager. &\multirow{2}{*}{\makecell{with \boldblue{open("file") as f}:\\ ~~~~~\boldred{content = f.read()}}} & /\\
\hline
children:parent$\rightarrow$child & 652417 & 569499 & Any pair of AST nodes that are parent and child in the parse tree. & / & / \\
\hline
comprehension:target$\rightarrow$iter & 3881 & / & A for clause to iterate over some sequences. & [x for \boldblue{x} in \boldred{list1}] & /\\
\bottomrule
\end{tabular}
}
\caption{Full dataset statistics table of \ours{}. For each relation type, we highlight the head and dependent nodes in the examples in bold, with the head in blue and the dependent in red. If a node can be either head or dependent in different relations, we color it in green. For more explanation about the syntax, please refer to the documentation of Python ast module~\cite{pythonast2021} and Java org.eclipse.jdt.core.dom.ASTParser~\cite{javaast2014}}
\label{tab:data_stats_full}
\end{table*}

\section{More Details on Attention Map Extraction for Code Language Models}
\label{app:attention-approach}

Our experiments follow the work of~\citet{clark2019does}. They evaluate the attention heads of BERT on dependency parsing tasks on an English dataset, while we extend the work to the PL domain. We adopt and extend some of their code, such as the functions for extracting attention from BERT and plotting attention weights. The main differences between our work and theirs are that we construct a novel dataset for syntax understanding tasks for PL and come up with related evaluation metrics to accommodate the characteristics of PL.

\subsection{Model Input}
Each of our code samples is an entire Python or Java function. To prepare the input to be fed to the models, we run CuBERT and CodeBERT tokenization to obtain sequences of input ids for each code sample. We insert a \texttt{[CLS]} token at the beginning and append a \texttt{[SEP]} token at the end. If the input length is longer than 512 tokens (the maximum number of tokens allowed), we discard that code sample. We never split a long code sample into several input sentences because the span of some dependency relations is very long within a function. For example, for an \texttt{if} statement, the \texttt{else} block may be far away from the keyword \texttt{if}. If we split them into two input sentences, then attention will not be able to understand and predict the relation between them. To avoid uncommon data points, we remove a code sample from both CuBERT and CodeBERT's input if it is longer than 512 tokens after either one of CuBERT or CodeBERT's tokenization.

\subsection{Token Alignment And Word-level Attention}
BERT uses WordPiece tokenization~\cite{DBLP:journals/corr/WuSCLNMKCGMKSJL16} and RoBERTa uses byte-level Byte-Pair Encoding (BPE)~\cite{DBLP:conf/acl/SennrichHB16a}, which may split a word into several subtokens. Additionally, CuBERT imposes some special rules when producing program vocabulary. However, our dataset's labels use code tokens generated by the tokenize module and the javalang module. Therefore, there exists a need to align CuBERT/CodeBERT subtokens with code tokens in order to evaluate the models on our dataset. We first generate such an alignment that maps each code token to a set of  CuBERT/CodeBERT subtokens, and then convert the original subtoken-level attention to word-level attention. We follow~\cite{clark2019does} to combine the attention weights of subtokens, i.e., we sum up their attention weights.

\section{More Reproducibility Information}
Here we provide more information according to the EMNLP 2022 Reproducibility Criteria.
\begin{itemize}
    \item Train/validation/test splits for datasets used: We do not finetune the pre-trained models on our benchmark. The validation set of \ours{}  contains the code samples that come from the validation set of CodeSearchNet, and our test set contains the samples from CodeSearchNet's test set. We use our test partition to probe the pre-trained attention heads while the validation set is not used.
    \item Number of parameters in each model: Cubert and BERT-large have 340M parameters. CodeBert and RoBERTa-base have 125M parameters. XLM-RoBERTa-base has 250M parameters. Multilingual BERT-base has 110M parameters.
    \item The average runtime for each model or algorithm: Running the pipeline to construct the \ours{} dataset takes about four hours assuming that dependencies and required datasets have been downloaded. The algorithm to probe a pre-trained model on one programming language of \ours{} takes about twelve hours on our machine using one Nvidia 1080Ti GPU.
\end{itemize}

\section{More Results on Top-k Scores}
\label{app:topk}

PL top-k scores are plotted in figure \ref{fig:PL_topk}. NL scores are plotted in figure \ref{fig:english} (English) and figure \ref{fig:german} (German).

\section{Examples of Correct and Incorrect Predictions}
\label{app:predictions}

In this section, we present some visualization examples where attention correctly or incorrectly predicts the dependents. The heads chosen in these examples are the best-performing heads of CuBERT evaluated using the first-token metric. We feed the entire function as input to the transformer, however, we only present relevant snippets here for simplicity. In the source code displayed, "..." denotes that the remaining part of the code is omitted. As a result, the attention from a token may not sum up to one in these figures because the rest of the function is omitted.

\begin{figure}[htbp]
\centering
\includegraphics[width=\linewidth]{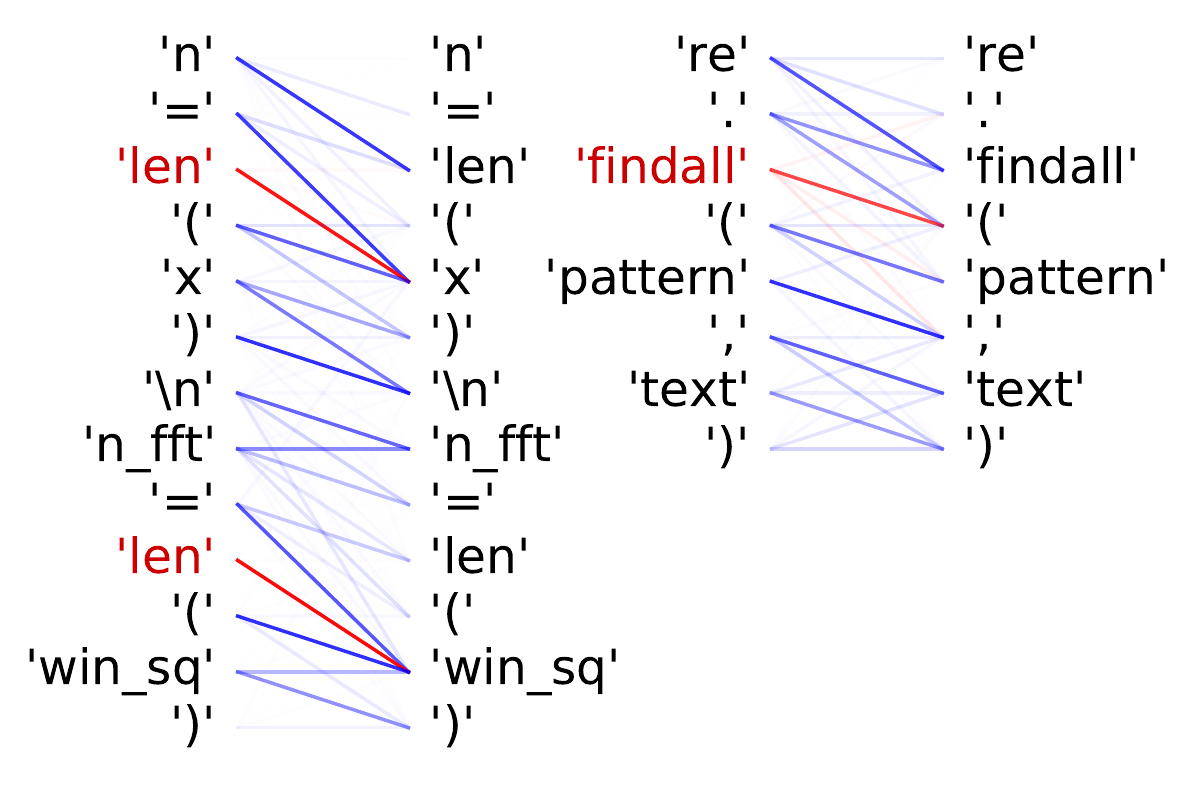}
\vspace{-1em}
\caption{Python Head 15-11 Call: func $\rightarrow$ args.}\label{case_func_arg}
\vspace{-1em}
\end{figure}

\begin{figure}[htbp]
\centering
\includegraphics[width=\linewidth]{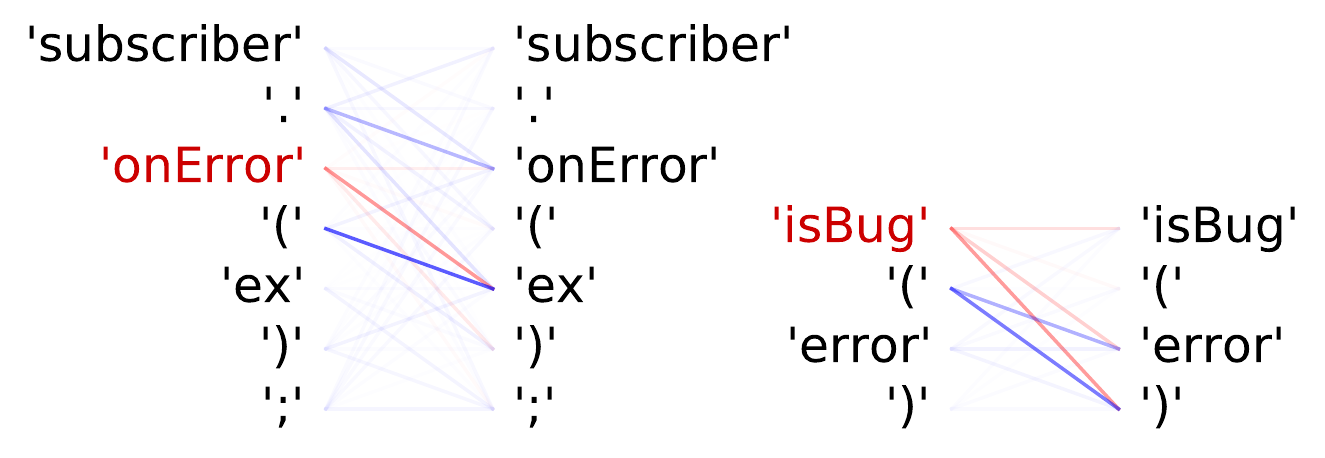}
\vspace{-1em}
\caption{Java Head 19-9 Call: func $\rightarrow$ args}\label{case_func_arg_java}
\vspace{-1em}
\end{figure}

\paragraph{Relation \texttt{Call: func $\rightarrow$ args}.}
The corresponding attention weights are visualized in Table~\ref{case_func_arg} for Python and \ref{case_func_arg_java} for Java.
\begin{itemize}[noitemsep, nolistsep]
\item Python correct case.
\begin{tcolorbox}[colback=yellow!15]
\vspace{-0.5em}
\begin{verbatim}
n = len(x)
n_fft = len(win_sq)
\end{verbatim}
\vspace{-0.5em}
\end{tcolorbox}
Attention correctly predicts the arguments \texttt{x} and \texttt{win\_sq}, respectively.

\item Python error case.
\begin{tcolorbox}[colback=yellow!15]
\vspace{-0.5em}
\begin{verbatim}
re.findall(pattern,text)
\end{verbatim}
\vspace{-0.5em}
\end{tcolorbox}
The function \texttt{findall} is called. The correct prediction should be the first argument, which is \texttt{pattern}; however, attention incorrectly predicts the parenthesis \texttt{(}.

\item Java correct case.
\begin{tcolorbox}[colback=yellow!15]
\vspace{-0.5em}
\begin{verbatim}
subscriber.onError(ex);
\end{verbatim}
\vspace{-0.5em}
\end{tcolorbox}
The token \texttt{ex} has the largest weight in attention, which is a correct prediction.

\item Java error case.
\begin{tcolorbox}[colback=yellow!15]
\vspace{-0.5em}
\begin{verbatim}
isBug(error)
\end{verbatim}
\vspace{-0.5em}
\end{tcolorbox}
The function \texttt{isBug} is called. The correct prediction should be the  argument, \texttt{error}; however, attention incorrectly predicts \texttt{)}.
\end{itemize}

\begin{figure}[htbp]
\centering
\includegraphics[width=\linewidth]{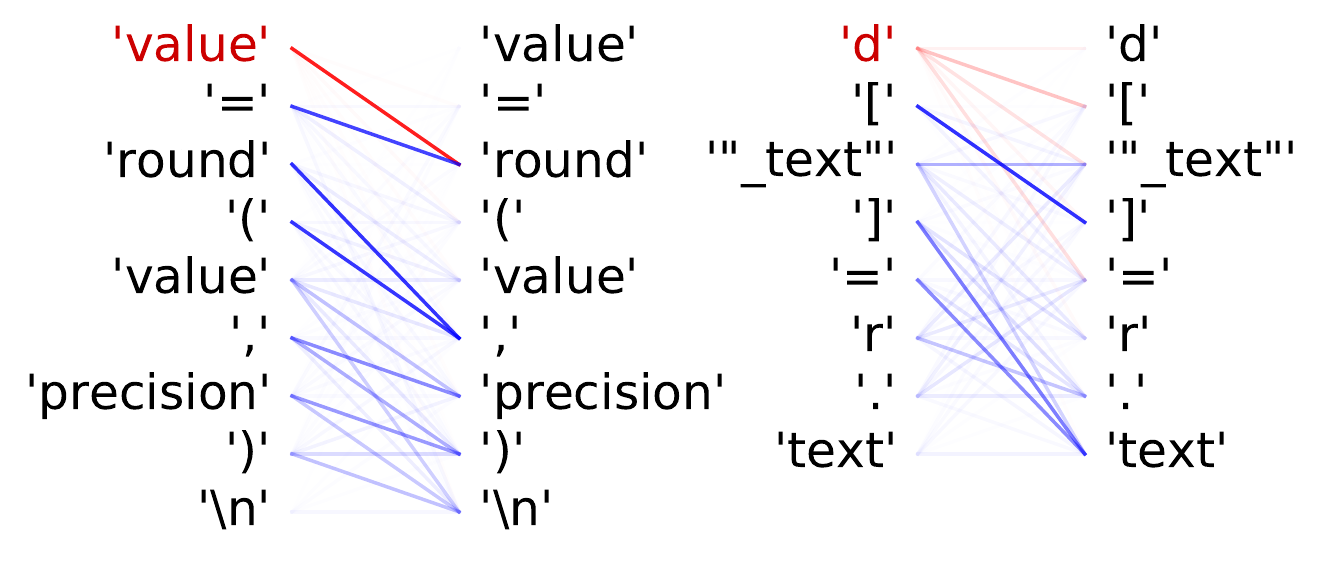}
\vspace{-1em}
\caption{Python Head 15-10 Assign: target $\rightarrow$ value}\label{case_assign}
\vspace{-1em}
\end{figure}

\begin{figure}[htbp]
\centering
\includegraphics[width=\linewidth]{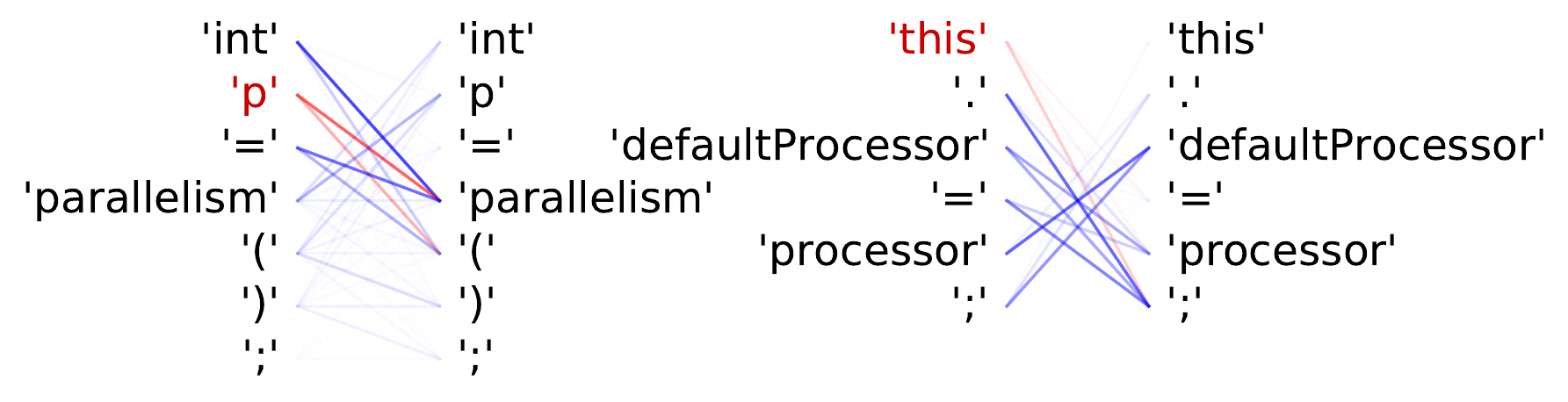}
\vspace{-1em}
\caption{Java Head 20-10 Assign: target $\rightarrow$ value}\label{case_assign_java}
\vspace{-1em}
\end{figure}

\paragraph{Relation \texttt{Assign: target $\rightarrow$ value}.}
The corresponding attention weights are visualized in Table~\ref{case_assign} for Python and \ref{case_assign_java} for Java.
\begin{itemize}[noitemsep, nolistsep]
\item Python correct case.
\begin{tcolorbox}[colback=yellow!15]
\vspace{-0.5em}
\begin{verbatim}
value = round(value, 
                precision)
\end{verbatim}
\vspace{-0.5em}
\end{tcolorbox}
The assigned value \texttt{round} is correctly predicted.

\item Python error case.
\begin{tcolorbox}[colback=yellow!15]
\vspace{-0.5em}
\begin{verbatim}
d["_text"] = r.text
\end{verbatim}
\vspace{-0.5em}
\end{tcolorbox}
The value assigned is \texttt{r.text}, but attention incorrectly predicts \texttt{[}.

\item Java correct case.
\begin{tcolorbox}[colback=yellow!15]
\vspace{-0.5em}
\begin{verbatim}
int p = parallelism();
\end{verbatim}
\vspace{-0.5em}
\end{tcolorbox}
Attention has the largest weight for the head token \texttt{parallelism}, which correctly predicts the relation.

\item Java error case.
\begin{tcolorbox}[colback=yellow!15]
\vspace{-0.5em}
\begin{verbatim}
this.defaultProcessor 
             = processor;
\end{verbatim}
\vspace{-0.5em}
\end{tcolorbox}
The value assigned is \texttt{processor}, but attention incorrectly predicts \texttt{;}.
\end{itemize}

\begin{figure}[htbp]
\centering
\includegraphics[width=\linewidth]{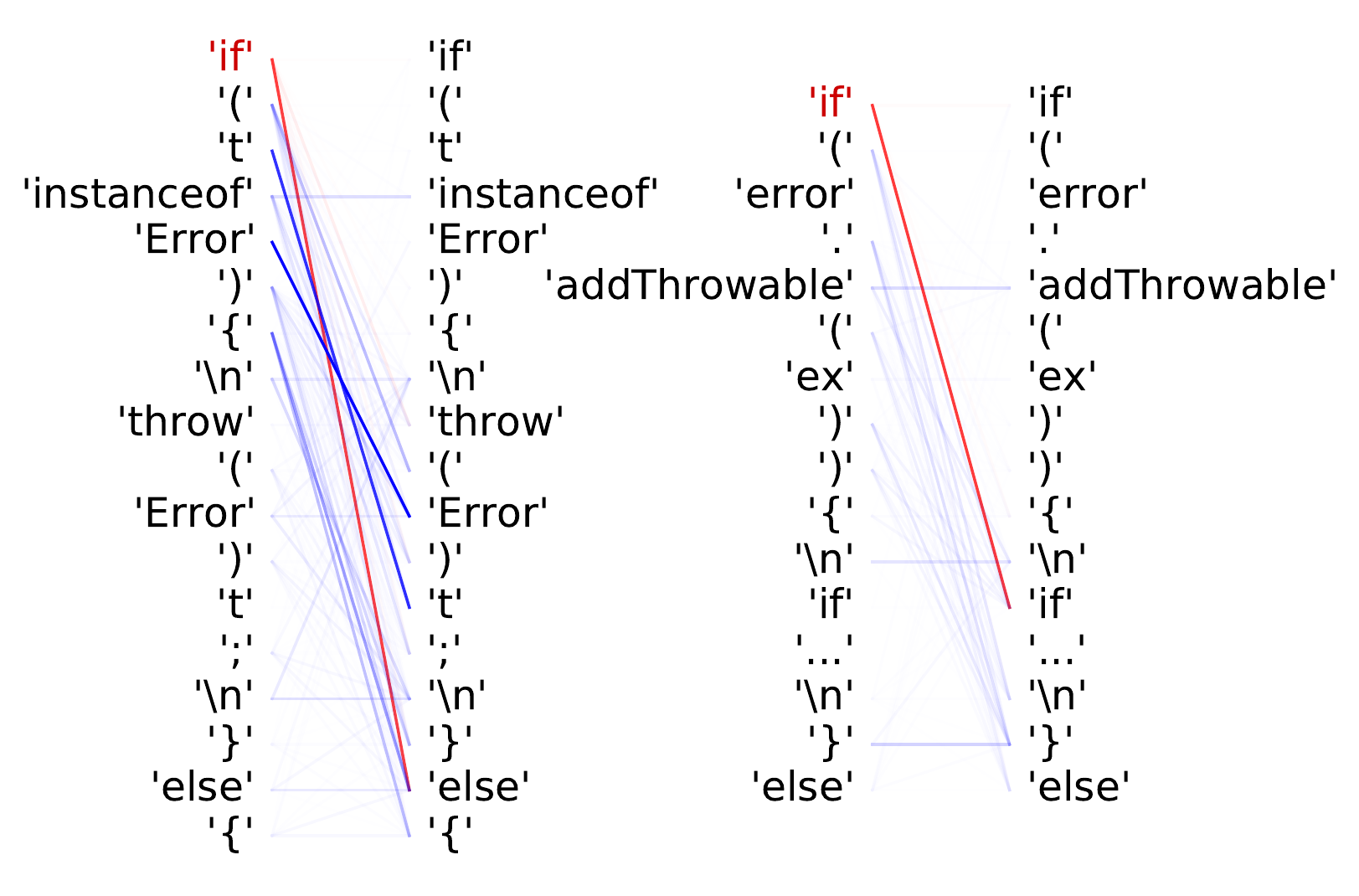}
\vspace{-1em}
\caption{Java Head 9-10 If: if $\rightarrow$ else}\label{case_if_else_java}
\vspace{-1em}
\end{figure}

\paragraph{Relation \texttt{If: if $\rightarrow$ else}.}
The corresponding attention weights are visualized in Table~\ref{case_if_else_java} for Java.
\begin{itemize}[noitemsep, nolistsep]
\item Java correct case.
\begin{tcolorbox}[colback=yellow!15]
\vspace{-0.5em}
\begin{verbatim}
if (t instanceof Error) {
    throw (Error) t;
} else { 
    ...
\end{verbatim}
\vspace{-0.5em}
\end{tcolorbox}
It correctly identifies the keyword \texttt{else}.

\item Java error case.
\begin{tcolorbox}[colback=yellow!15]
\vspace{-0.5em}
\begin{verbatim}
if(error.addThrowable(ex)) {
    if ...
} else {
    ...
\end{verbatim}
\vspace{-0.5em}
\end{tcolorbox}
There is another \texttt{if} statement inside the body of the first if statement. The correct prediction should be keyword \texttt{else}, but it predicts the inner \texttt{if}.
\end{itemize}

\begin{figure}[htbp]
\centering
\includegraphics[width=\linewidth]{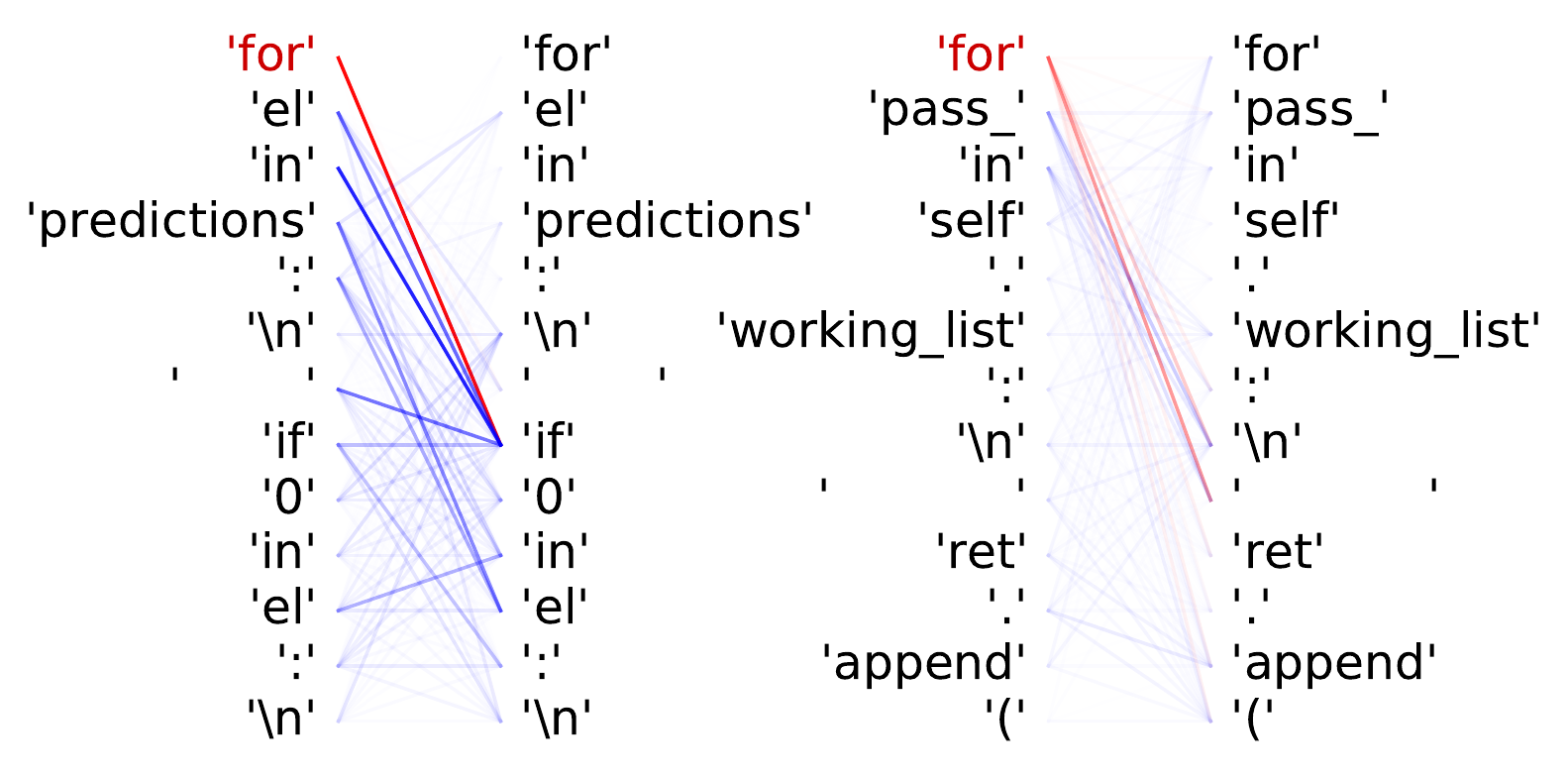}
\vspace{-1em}
\caption{Python Head 18-4 For: for $\rightarrow$ body}\label{case_for_body}
\vspace{-1em}
\end{figure}

\begin{figure}[htbp]
\centering
\includegraphics[width=\linewidth]{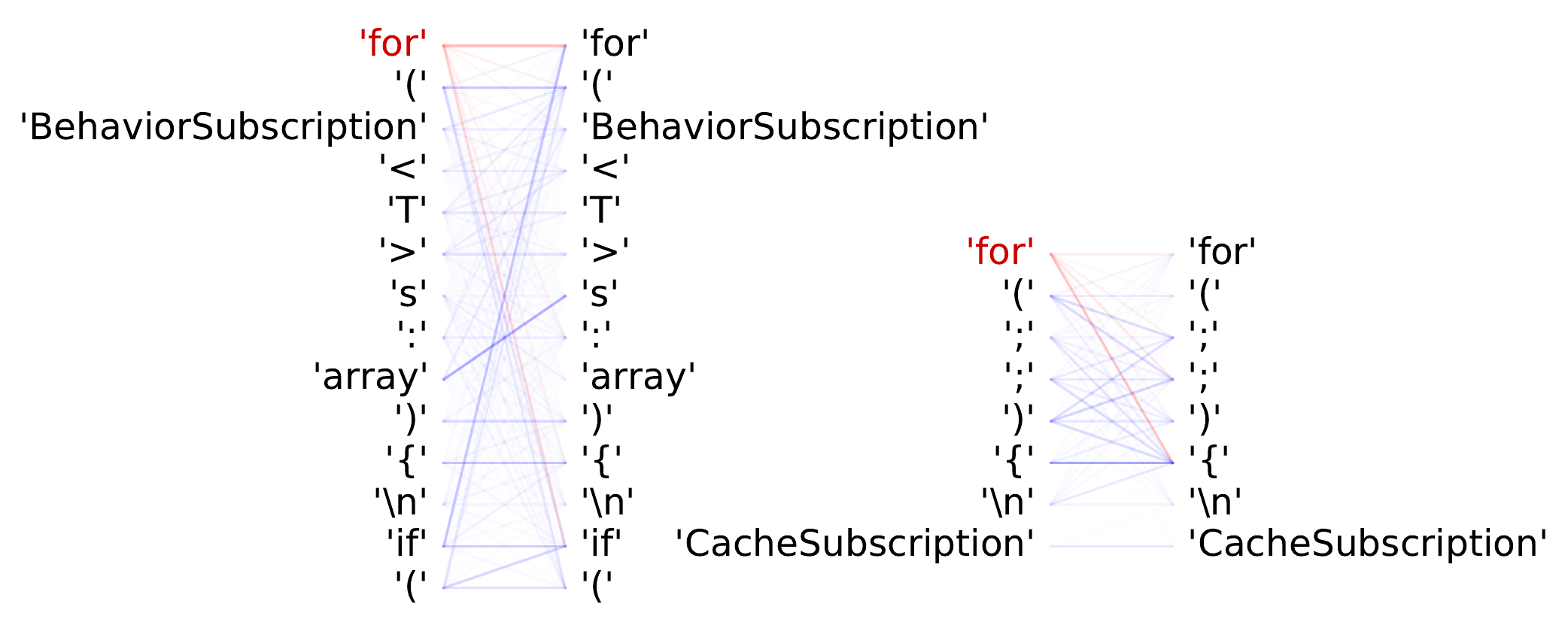}
\vspace{-1em}
\caption{Java Head 16-5 For: for $\rightarrow$ body}\label{case_for_body_java}
\vspace{-1em}
\end{figure}

\paragraph{Relation \texttt{For: for $\rightarrow$ body}.}
The corresponding attention weights are visualized in Table~\ref{case_for_body} for Python and \ref{case_for_body_java} for Java.
\begin{itemize}[noitemsep, nolistsep]
\item Python correct case.
\begin{tcolorbox}[colback=yellow!15]
\vspace{-0.5em}
\begin{verbatim}
for el in predictions:
    if 0 in el:
        ...
\end{verbatim}
\vspace{-0.5em}
\end{tcolorbox}

\item Python error case.
\begin{tcolorbox}[colback=yellow!15]
\vspace{-0.5em}
\begin{verbatim}
for pass_ in 
        self.working_list:
    ret.append(...
\end{verbatim}
\vspace{-0.5em}
\end{tcolorbox}
The correct prediction should be the first token within the body, which is \texttt{ret}; however, attention incorrectly predicts the blank space "~~~~" before \texttt{ret}.

\item Java correct case.
\begin{tcolorbox}[colback=yellow!15]
\vspace{-0.5em}
\begin{verbatim}
for(BehaviorSubscription<T> 
               s : array) {
    if (...
\end{verbatim}
\vspace{-0.5em}
\end{tcolorbox}

\item Java error case.
\begin{tcolorbox}[colback=yellow!15]
\vspace{-0.5em}
\begin{verbatim}
for (;;) {
    CacheSubscription ...
\end{verbatim}
\vspace{-0.5em}
\end{tcolorbox}
The correct prediction should be the first token within the body, which is \texttt{CacheSubscription}; however, attention incorrectly predicts \texttt{\{}.
\end{itemize}

\section{More Results on the Ablation Study of Delimiter Tokens}
\label{app:separation-tokens}

\begin{figure}[htbp]
\centering
\subfigure[Original]{%
\includegraphics[width=\linewidth]{figures/java_test_last_semicolon_in_keywords.pdf}}
\subfigure[With Newline]{%
\includegraphics[width=\linewidth]{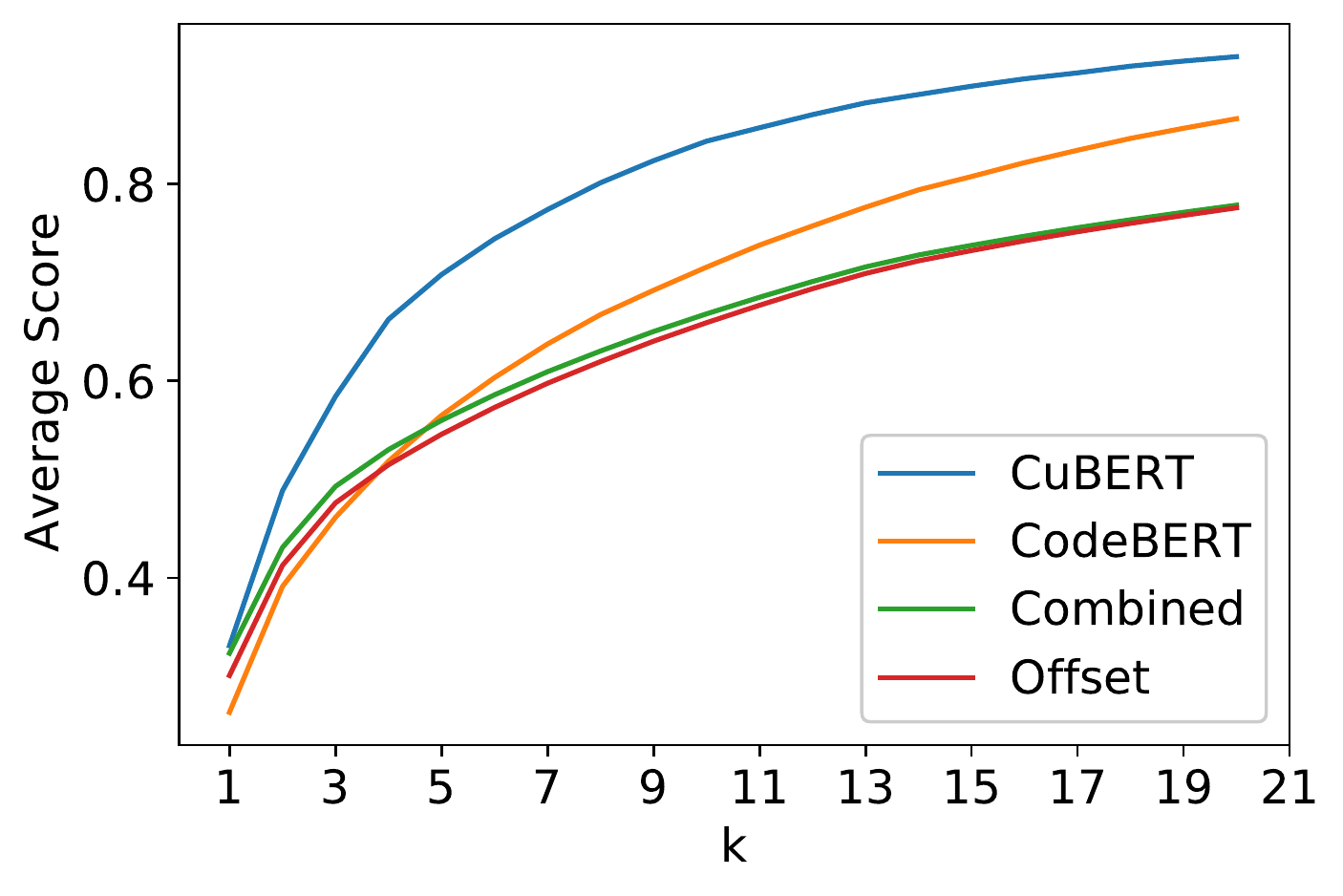}}
\subfigure[Without Semicolons]{%
\includegraphics[width=\linewidth]{figures/java_test_last_skip_semicolon.pdf}}
\caption{Top-k scores for Java syntax understanding using the last-token metric.}
\label{ablation_java_last}
\end{figure}

\begin{figure}[htbp]
\centering
\subfigure[Original]{%
\includegraphics[width=\linewidth]{figures/java_test_any_semicolon_in_keywords.pdf}} 
\subfigure[With Newline]{%
\includegraphics[width=\linewidth]{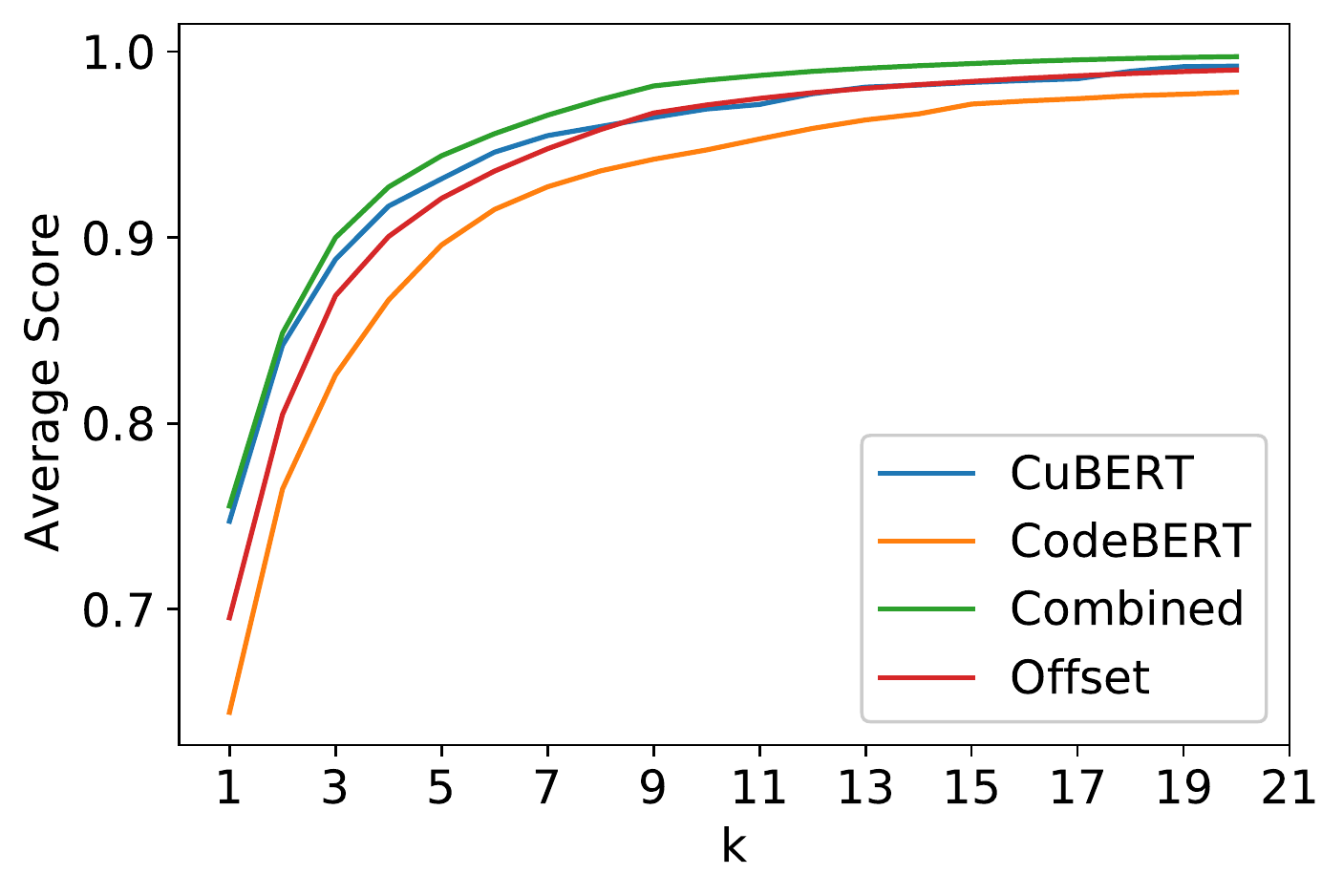}} 
\subfigure[Without Semicolons]{%
\includegraphics[width=\linewidth]{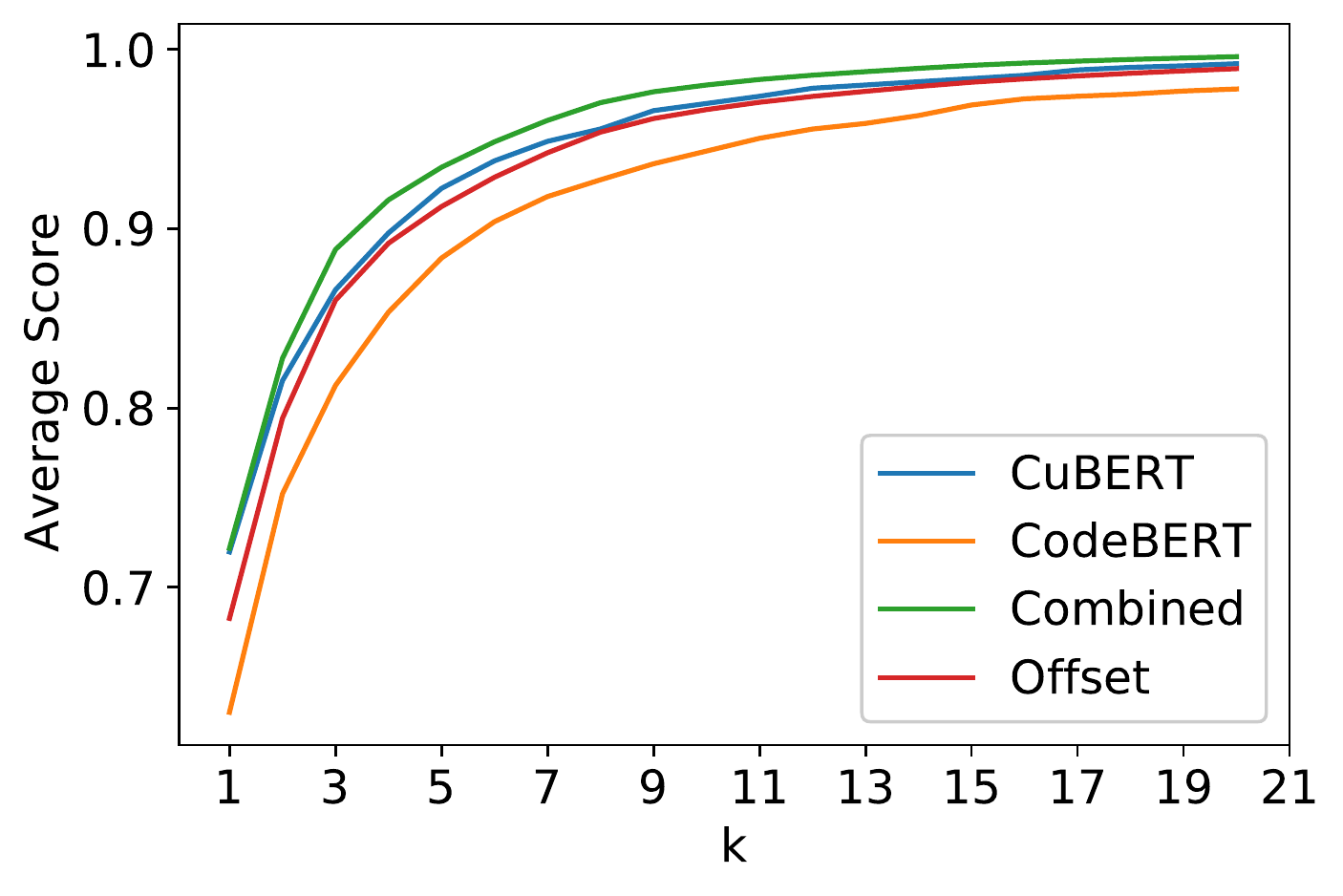}}
\caption{Ablation Study (Java Any-Token Metric).} \label{ablation_java_any}
\end{figure}

The ablation study on java dataset is shown in Figure~\ref{ablation_java_any} (any-token metric) and Figure~\ref{ablation_java_last} (last-token metric). Results with first-token metric are not affected at all because semicolons and newlines are never used as the first token of dependents. We found that attention performs very well with last-token metric because it can find semicolons and newlines.  

\begin{figure}[htbp]
\centering
\subfigure[Original]{%
\includegraphics[width=\linewidth]{figures/python_test_any.pdf}} 
\subfigure[With Newline]{%
\includegraphics[width=\linewidth]{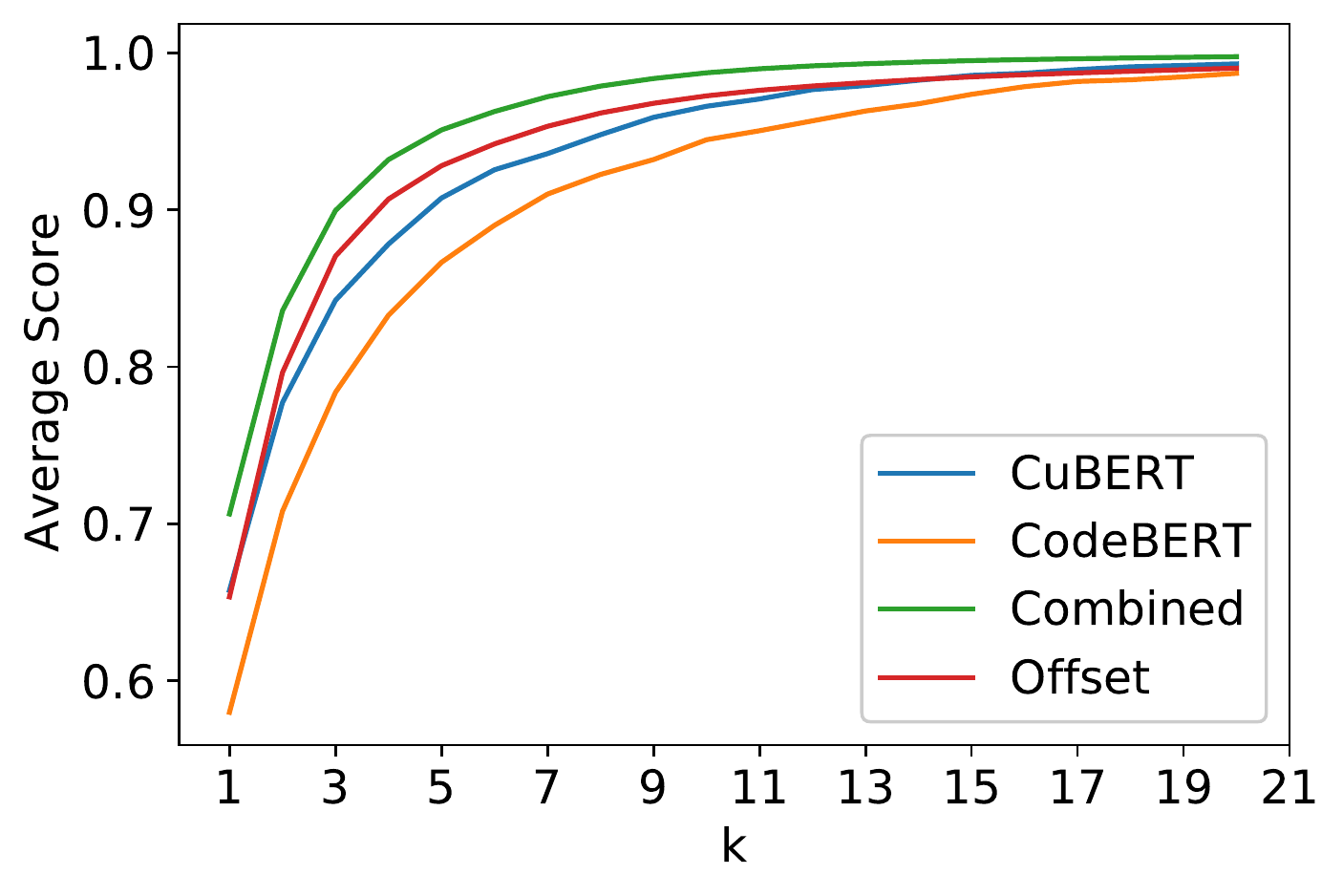}} 
\caption{Ablation Study (Python Any-Token Metric).}\label{ablation_python_any}
\end{figure}

\begin{figure}[htbp]
\centering
\subfigure[Original]{%
\includegraphics[width=\linewidth]{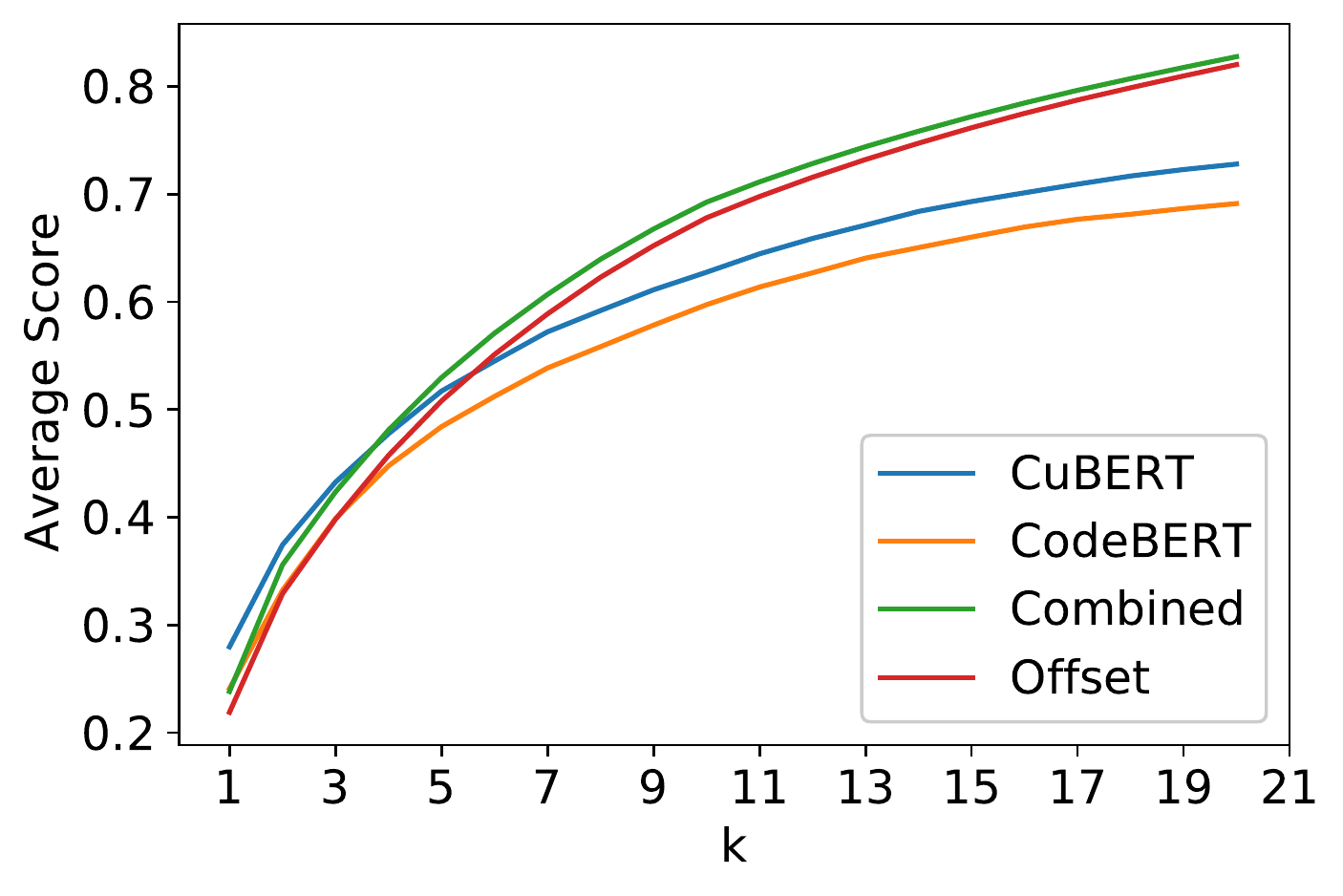}} 
\subfigure[With Newline]{%
\includegraphics[width=\linewidth]{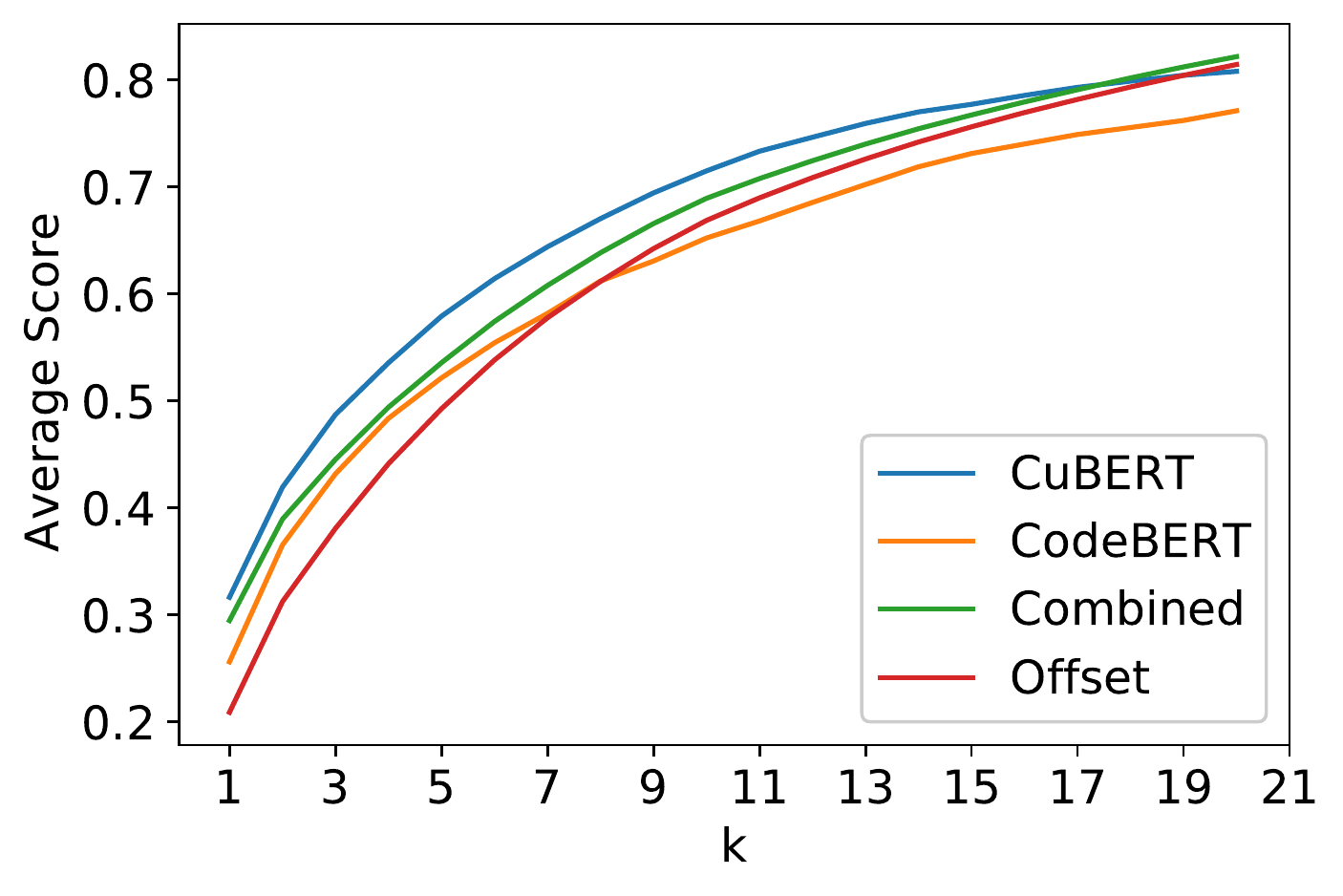}} 
\caption{Ablation Study (Python Last-Token Metric).}\label{ablation_python_last}
\end{figure}

The ablation study on python dataset is shown in Figure~\ref{ablation_python_any} (any-token metric) and Figure~\ref{ablation_python_last} (last-token metric).

\section{More Breakdown Results on Different Relation Types}
\label{app:relation-breakdown}
More results on comparisons  of  top-1   scores between  CuBERT  and  the  offset  baseline are presented in Tables~\ref{tab:attn_offset_python_first}, \ref{tab:attn_offset_python_any}, \ref{tab:attn_offset_java_first}, and \ref{tab:attn_offset_java_any}.

\begin{table*}[ht]
\begin{tabular}{@{}lllll@{}}
\toprule
 \multirow{2}{*}{Relation} & \multicolumn{2}{c}{Score} & \multirow{2}{*}{Offset} & \multirow{2}{*}{Difference} \\
     &     CuBERT      &     Offset     &  &      \\ \midrule
If:if$\rightarrow$else & 92.7 & 5.7 & 17 & 87.1 \\
IfExp:body$\rightarrow$orelse & 46.4 & 13.6 & 6 & 32.8 \\
Try:body$\rightarrow$handler & 39.1 & 7.9 & 8 & 31.3 \\
If:body$\rightarrow$orelse & 29.2 & 7.1 & 12 & 22.0 \\
BoolOp:value$\rightarrow$value & 33.3 & 22.4 & 2 & 10.9 \\
Try:body$\rightarrow$finalbody & 20.5 & 10.3 & 7 & 10.3 \\
If:if$\rightarrow$body & 31.5 & 23.1 & 7 & 8.4 \\
Call:func$\rightarrow$keywords & 38.0 & 34.1 & 2 & 4.0 \\
If:test$\rightarrow$orelse & 7.5 & 5.1 & 18 & 2.4 \\
Compare:left$\rightarrow$comparator & 46.5 & 45.7 & 2 & 0.8 \\
If:if$\rightarrow$test & 98.8 & 98.8 & 1 & -0.0 \\
For:for$\rightarrow$target & 99.4 & 99.4 & 1 & 0.0 \\
While:while$\rightarrow$test & 99.3 & 99.3 & 1 & 0.0 \\
Try:body$\rightarrow$orelse & 10.5 & 10.5 & 33 & 0.0 \\
For:for$\rightarrow$body & 30.4 & 32.7 & 7 & -2.3 \\
Try:handler$\rightarrow$orelse & 13.2 & 15.8 & 16 & -2.6 \\
IfExp:test$\rightarrow$orelse & 23.2 & 26.7 & 2 & -3.5 \\
children:parent$\rightarrow$child & 26.3 & 30.7 & 2 & -4.4 \\
Attribute:value$\rightarrow$attr & 76.7 & 82.3 & 2 & -5.6 \\
For:target$\rightarrow$body & 26.4 & 32.7 & 6 & -6.3 \\
If:test$\rightarrow$body & 16.3 & 23.2 & 6 & -6.9 \\
Subscript:value$\rightarrow$slice & 59.4 & 66.4 & 2 & -7.1 \\
BinOp:left$\rightarrow$right & 31.5 & 45.2 & 2 & -13.7 \\
Try:handler$\rightarrow$finalbody & 8.3 & 25.0 & 21 & -16.7 \\
For:target$\rightarrow$iter & 58.3 & 77.4 & 2 & -19.1 \\
AugAssign:target$\rightarrow$value & 49.4 & 70.3 & 2 & -21.0 \\
For:iter$\rightarrow$body & 11.7 & 34.8 & 4 & -23.1 \\
IfExp:body$\rightarrow$test & 17.2 & 42.1 & 2 & -24.9 \\
While:while$\rightarrow$body & 22.1 & 48.5 & 5 & -26.5 \\
Assign:target$\rightarrow$value & 39.8 & 71.2 & 2 & -31.4 \\
While:test$\rightarrow$body & 16.2 & 48.5 & 4 & -32.4 \\
Call:func$\rightarrow$args & 59.3 & 93.2 & 2 & -33.9 \\
Call:args$\rightarrow$keywords & 20.6 & 54.9 & 2 & -34.4 \\
For:for$\rightarrow$iter & 34.1 & 77.4 & 3 & -43.3 \\
 \bottomrule
\end{tabular}
\caption{Attention vs. offset baseline with fixed offset for each relation on \textbf{Python} dataset using \textbf{first}-token metric. In the score column, we present the accuracy score for CuBERT and offset baseline. In the offset column, the chosen offset is shown. Score differences are calculated as CuBERT score - offset baseline score for each relation,  where  a  positive  value  indicates  that  the  language  model  surpasses  the  baseline performance.  Since CuBERT always outperforms Codebert, we only include results for CuBERT.}
\label{tab:attn_offset_python_first}
\end{table*}

\begin{table*}[ht]
\begin{tabular}{@{}lllll@{}}
\toprule
 \multirow{2}{*}{Relation} & \multicolumn{2}{c}{Score} & \multirow{2}{*}{Offset} & \multirow{2}{*}{Difference} \\
     &     CuBERT      &     Offset     &  &      \\ \midrule
If:if$\rightarrow$else & 92.7 & 5.7 & 17 & 87.1 \\
IfExp:body$\rightarrow$orelse & 52.4 & 21.3 & 10 & 31.1 \\
For:target$\rightarrow$iter & 98.8 & 77.4 & 2 & 21.4 \\
For:target$\rightarrow$body & 97.2 & 81.4 & 14 & 15.8 \\
If:if$\rightarrow$body & 77.0 & 61.8 & 11 & 15.2 \\
If:body$\rightarrow$orelse & 60.7 & 48.7 & 18 & 12.0 \\
For:for$\rightarrow$body & 93.2 & 81.3 & 15 & 11.8 \\
Compare:left$\rightarrow$comparator & 56.4 & 45.7 & 2 & 10.7 \\
Try:body$\rightarrow$orelse & 57.9 & 52.6 & 62 & 5.3 \\
While:while$\rightarrow$body & 95.6 & 92.6 & 14 & 2.9 \\
children:parent$\rightarrow$child & 45.3 & 42.6 & 2 & 2.7 \\
If:if$\rightarrow$test & 100.0 & 98.8 & 1 & 1.1 \\
BinOp:left$\rightarrow$right & 46.0 & 45.2 & 2 & 0.8 \\
For:for$\rightarrow$target & 99.4 & 99.4 & 1 & 0.0 \\
While:while$\rightarrow$test & 99.3 & 99.3 & 1 & 0.0 \\
Try:body$\rightarrow$finalbody & 25.6 & 25.6 & 56 & 0.0 \\
Assign:target$\rightarrow$value & 78.2 & 79.7 & 4 & -1.5 \\
Call:func$\rightarrow$keywords & 62.4 & 64.7 & 4 & -2.3 \\
If:test$\rightarrow$body & 59.2 & 61.7 & 10 & -2.5 \\
BoolOp:value$\rightarrow$value & 47.8 & 50.6 & 8 & -2.8 \\
AugAssign:target$\rightarrow$value & 66.8 & 70.3 & 2 & -3.6 \\
Subscript:value$\rightarrow$slice & 61.7 & 66.4 & 2 & -4.8 \\
For:for$\rightarrow$iter & 72.2 & 77.4 & 3 & -5.2 \\
Attribute:value$\rightarrow$attr & 76.7 & 82.3 & 2 & -5.6 \\
IfExp:body$\rightarrow$test & 34.8 & 42.1 & 2 & -7.3 \\
While:test$\rightarrow$body & 85.3 & 92.6 & 13 & -7.4 \\
If:test$\rightarrow$orelse & 36.5 & 44.9 & 31 & -8.4 \\
Try:body$\rightarrow$handler & 41.6 & 51.2 & 17 & -9.7 \\
IfExp:test$\rightarrow$orelse & 30.0 & 40.7 & 4 & -10.7 \\
Try:handler$\rightarrow$orelse & 31.6 & 47.4 & 25 & -15.8 \\
Call:func$\rightarrow$args & 75.3 & 93.2 & 2 & -17.8 \\
For:iter$\rightarrow$body & 63.6 & 83.5 & 12 & -19.8 \\
Call:args$\rightarrow$keywords & 49.2 & 74.4 & 4 & -25.2 \\
Try:handler$\rightarrow$finalbody & 16.7 & 58.3 & 21 & -41.7 \\
 \bottomrule
\end{tabular}
\caption{Attention vs. offset baseline with fixed offset for each relation on \textbf{Python} dataset using \textbf{any}-token metric.}
\label{tab:attn_offset_python_any}
\end{table*}

\begin{table*}[ht]
\begin{tabular}{@{}lllll@{}}
\toprule
 \multirow{2}{*}{Relation} & \multicolumn{2}{c}{Score} & \multirow{2}{*}{Offset} & \multirow{2}{*}{Difference} \\
     &     CuBERT      &     Offset     &  &      \\ \midrule
If:if$\rightarrow$else & 87.0 & 7.0 & 15 & 80.0 \\
Switch:switch$\rightarrow$statement & 100.0 & 75.2 & 5 & 24.8 \\
If:if$\rightarrow$body & 48.8 & 26.2 & 7 & 22.5 \\
For:test$\rightarrow$updaters & 76.0 & 53.4 & 4 & 22.5 \\
If:body$\rightarrow$orelse & 28.7 & 9.4 & 10 & 19.3 \\
Try:body$\rightarrow$handler & 24.4 & 5.5 & 8 & 18.9 \\
Do:body$\rightarrow$test & 13.3 & 6.7 & 74 & 6.7 \\
Try:body$\rightarrow$finalbody & 7.0 & 4.3 & 9 & 2.6 \\
Do:do$\rightarrow$test & 6.7 & 6.7 & 34 & 0.0 \\
IfExp:test$\rightarrow$orelse & 24.3 & 24.5 & 7 & -0.2 \\
InstanceofExpr:expr$\rightarrow$type & 89.8 & 91.9 & 2 & -2.1 \\
For:for$\rightarrow$initializers & 97.5 & 100.0 & 2 & -2.5 \\
Attribute:value$\rightarrow$attr & 81.2 & 83.9 & 2 & -2.7 \\
children:parent$\rightarrow$child & 34.2 & 36.8 & 2 & -2.7 \\
If:test$\rightarrow$orelse & 2.6 & 6.8 & 15 & -4.2 \\
For:initializers$\rightarrow$updaters & 39.8 & 45.7 & 9 & -6.0 \\
Subscript:value$\rightarrow$slice & 72.3 & 78.8 & 2 & -6.5 \\
IfExp:body$\rightarrow$orelse & 44.2 & 52.5 & 2 & -8.2 \\
For:for$\rightarrow$updaters & 36.2 & 45.5 & 11 & -9.3 \\
Try:handler$\rightarrow$finalbody & 19.5 & 29.9 & 16 & -10.4 \\
InfixExpr:left$\rightarrow$right & 50.7 & 61.6 & 2 & -10.9 \\
Switch:switch$\rightarrow$expr & 89.0 & 100.0 & 2 & -11.0 \\
If:test$\rightarrow$body & 11.3 & 26.2 & 5 & -14.9 \\
IfExp:test$\rightarrow$body & 21.2 & 37.5 & 5 & -16.3 \\
For:initializers$\rightarrow$body & 21.0 & 37.5 & 13 & -16.6 \\
Switch:expr$\rightarrow$statement & 58.1 & 75.2 & 3 & -17.1 \\
For:for$\rightarrow$body & 16.1 & 36.0 & 15 & -19.9 \\
While:while$\rightarrow$body & 13.2 & 42.7 & 9 & -29.6 \\
While:test$\rightarrow$body & 9.4 & 42.7 & 7 & -33.3 \\
Assign:target$\rightarrow$value & 35.3 & 68.7 & 2 & -33.4 \\
Call:func$\rightarrow$args & 63.6 & 98.7 & 2 & -35.1 \\
For:test$\rightarrow$body & 13.9 & 49.2 & 8 & -35.3 \\
If:if$\rightarrow$test & 58.2 & 96.5 & 2 & -38.3 \\
While:while$\rightarrow$test & 41.9 & 82.0 & 2 & -40.1 \\
For:updaters$\rightarrow$body & 19.4 & 88.9 & 4 & -69.5 \\
For:initializers$\rightarrow$test & 15.1 & 85.4 & 5 & -70.3 \\
Do:do$\rightarrow$body & 26.7 & 100.0 & 2 & -73.3 \\
For:for$\rightarrow$test & 10.5 & 84.8 & 7 & -74.3 \\
 \bottomrule
\end{tabular}
\caption{Attention vs. offset baseline with fixed offset for each relation on \textbf{Java} dataset using \textbf{first}-token metric.}
\label{tab:attn_offset_java_first}
\end{table*}

\begin{table*}[ht]
\begin{tabular}{@{}lllll@{}}
\toprule
 \multirow{2}{*}{Relation} & \multicolumn{2}{c}{Score} & \multirow{2}{*}{Offset} & \multirow{2}{*}{Difference} \\
     &     CuBERT      &     Offset     &  &      \\ \midrule
If:if$\rightarrow$else & 87.0 & 7.0 & 15 & 80.0 \\
For:test$\rightarrow$updaters & 86.9 & 53.6 & 5 & 33.3 \\
If:if$\rightarrow$body & 79.6 & 62.1 & 11 & 17.5 \\
If:body$\rightarrow$orelse & 69.1 & 52.1 & 16 & 17.0 \\
While:while$\rightarrow$test & 98.8 & 82.0 & 2 & 16.9 \\
Assign:target$\rightarrow$value & 81.4 & 68.7 & 2 & 12.6 \\
Try:body$\rightarrow$finalbody & 20.0 & 13.0 & 37 & 7.0 \\
IfExp:body$\rightarrow$orelse & 59.5 & 52.5 & 2 & 7.0 \\
For:for$\rightarrow$updaters & 53.6 & 46.8 & 11 & 6.7 \\
Do:body$\rightarrow$test & 33.3 & 26.7 & 81 & 6.7 \\
Do:do$\rightarrow$test & 33.3 & 26.7 & 81 & 6.7 \\
For:for$\rightarrow$test & 95.3 & 89.5 & 9 & 5.8 \\
IfExp:test$\rightarrow$orelse & 45.9 & 42.4 & 10 & 3.5 \\
While:while$\rightarrow$body & 91.4 & 89.0 & 17 & 2.4 \\
If:if$\rightarrow$test & 98.7 & 96.5 & 2 & 2.3 \\
children:parent$\rightarrow$child & 44.0 & 42.4 & 2 & 1.6 \\
InstanceofExpr:expr$\rightarrow$type & 93.1 & 91.9 & 2 & 1.1 \\
For:initializers$\rightarrow$updaters & 47.2 & 47.1 & 9 & 0.1 \\
Switch:switch$\rightarrow$statement & 100.0 & 100.0 & 21 & 0.0 \\
For:for$\rightarrow$initializers & 99.4 & 100.0 & 2 & -0.6 \\
Try:body$\rightarrow$handler & 44.0 & 44.7 & 25 & -0.7 \\
Subscript:value$\rightarrow$slice & 77.9 & 78.8 & 2 & -0.9 \\
Switch:switch$\rightarrow$expr & 99.0 & 100.0 & 2 & -1.0 \\
While:test$\rightarrow$body & 87.4 & 89.5 & 15 & -2.2 \\
Attribute:value$\rightarrow$attr & 81.2 & 83.9 & 2 & -2.7 \\
For:updaters$\rightarrow$body & 96.8 & 99.6 & 9 & -2.8 \\
Switch:expr$\rightarrow$statement & 96.6 & 100.0 & 12 & -3.4 \\
InfixExpr:left$\rightarrow$right & 58.2 & 61.6 & 2 & -3.4 \\
For:test$\rightarrow$body & 92.3 & 96.1 & 14 & -3.9 \\
If:test$\rightarrow$body & 57.4 & 62.1 & 9 & -4.7 \\
For:for$\rightarrow$body & 90.3 & 94.9 & 23 & -4.7 \\
If:test$\rightarrow$orelse & 42.3 & 47.7 & 27 & -5.4 \\
Do:do$\rightarrow$body & 93.3 & 100.0 & 2 & -6.7 \\
For:initializers$\rightarrow$body & 87.7 & 94.9 & 21 & -7.2 \\
Call:func$\rightarrow$args & 91.1 & 98.7 & 2 & -7.6 \\
For:initializers$\rightarrow$test & 74.5 & 90.2 & 7 & -15.6 \\
IfExp:test$\rightarrow$body & 25.9 & 50.0 & 5 & -24.1 \\
Try:handler$\rightarrow$finalbody & 26.0 & 53.2 & 20 & -27.3 \\
 \bottomrule
\end{tabular}
\caption{Attention vs. offset baseline with fixed offset for each relation on \textbf{Java} dataset using \textbf{any}-token metric.}
\label{tab:attn_offset_java_any}
\end{table*}

\end{document}